%% file: main.tex
\pdfoutput=1
\documentclass{article}
\usepackage{iclr2027_conference,times}
\iclrfinalcopy
\input{math_commands.tex}

\usepackage[utf8]{inputenc}
\usepackage[T1]{fontenc}
\usepackage{graphicx}
\usepackage{placeins}
\usepackage{booktabs}
\usepackage{amsmath}
\usepackage{amssymb}
\usepackage{amsfonts}
\usepackage{nicefrac}
\usepackage{xcolor}
\usepackage{colortbl}
\usepackage{multirow}
\usepackage{rotating}
\usepackage{enumitem}
\usepackage{subcaption}
\usepackage{listings}
\usepackage{wrapfig}
\usepackage{hyperref}
\usepackage{url}

\definecolor{codegreen}{RGB}{45,120,120}
\definecolor{codepink}{RGB}{220,20,120}

\lstdefinestyle{pytorchstyle}{
    language=Python,
    basicstyle=\ttfamily\footnotesize,
    numbers=left,
    numberstyle=\tiny\color{gray},
    numbersep=6pt,
    xleftmargin=12pt,
    keywordstyle=\color{codepink}\bfseries,
    commentstyle=\color{codegreen},
    stringstyle=\color{purple},
    showstringspaces=false,
    breaklines=true,
    frame=lines,
    columns=fullflexible,
    keepspaces=true,
    morekeywords={for,if,else,return,True,False,None,in,def,class,with,as,continue,is}
}

\title{DBCM: Dual-Branch Calibration Module\\
for Test-Time Adaptation in Time Series Forecasting}

\author{
  Jaebyeong Jeon$^{1}$ \qquad
  Kunwoo Kang$^{2}$ \qquad
  Sang Yup Lee$^{2,3}$ \qquad
  Kibok Lee$^{1}$ \\
  $^{1}$Department of Statistics and Data Science, Yonsei University \\
  $^{2}$Digital Analytics, College of Computing, Yonsei University \\
  $^{3}$Department of Communication, Yonsei University \\
  \texttt{\{jaebyeong98, kunwoo2027, sangyuplee, kibok\}@yonsei.ac.kr}
}

\begin{document}

\maketitle
\lhead{Preprint}
\vspace{-0.2in}

\begin{abstract}
Non-stationarity induces test-time distribution shifts in time series forecasting (TSF), resulting in forecast residuals between predictions and the corresponding ground truth.
Recent works on test-time adaptation for TSF address this by introducing lightweight calibration modules around frozen forecasters, updated using observations that become incrementally available at test time.
Existing approaches typically employ channel-wise adapters, implicitly assuming that forecast residuals are inherently channel-specific.
However, we observe that forecast residuals often exhibit substantial shared structure across channels, suggesting that fully independent channel-wise parameterization may be redundant.
Motivated by this observation, we propose the \textbf{D}ual-\textbf{B}ranch \textbf{C}alibration \textbf{M}odule (DBCM), which shares adaptation capacity across channels through a shared adapter while retaining limited channel-specific flexibility through a low-rank channel-wise adapter, further enhanced by channel mixing and nonlinear gated modulation.
Extensive experiments show that DBCM consistently outperforms prior methods across diverse datasets, horizons, and backbones.
\end{abstract}

\section{Introduction}
\label{sec:intro}

Time series forecasting (TSF) aims to predict future values based on historical observations, often involving multiple correlated variables. It serves as a fundamental tool in various real-world applications, including financial markets, energy systems, and weather forecasting~\citep{petropoulos2022forecasting, verma2024climode, liu2023spatio, chen2023financial, dipaizza2021energy}.
Despite its broad applicability, TSF is often challenged by non-stationarity, where test-time distribution shifts arise from evolving system dynamics or external factors~\citep{kim2021reversible, liu2022nonstationary, fan2023dish, liu2023adaptive, petropoulos2022forecasting}.
Such shifts degrade the performance of pre-trained forecasters, which are typically trained offline under the assumption that training and test data follow similar distributions.
To address this challenge, recent studies have adopted test-time adaptation (TTA), which dynamically adjusts a pre-trained model during deployment.
In computer vision, TTA commonly relies on entropy minimization over unlabeled test inputs~\citep{wang2020tent, niu2022efficient}.
In TSF, however, observations become available after prediction and can serve as direct supervision.
Accordingly, TSF-TTA frameworks leverage observed ground truth at test time to update lightweight calibration modules, while keeping the source forecaster frozen~\citep{kim2025battling, medeiros2025accurate, grover2025shiftaware, im2026cosa}, thereby improving forecasting performance under distribution shifts.

\begin{figure}[t]
    \centering
    \begin{subfigure}[t]{0.48\linewidth}
        \centering
        \includegraphics[width=\linewidth]{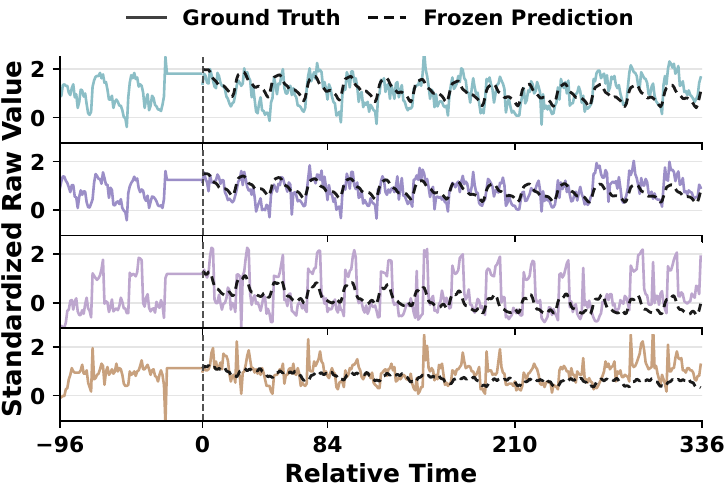}
        \caption{Raw channel behaviors.}
    \end{subfigure}
    \hfill
    \begin{subfigure}[t]{0.48\linewidth}
        \centering
        \includegraphics[width=\linewidth]{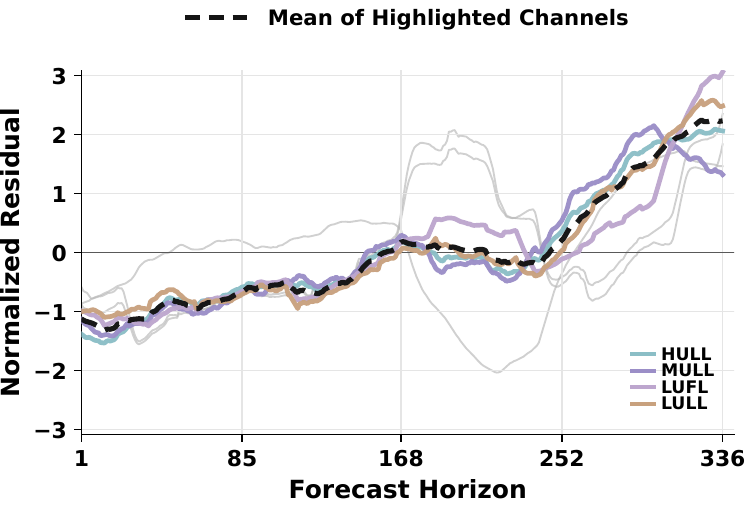}
        \caption{Residual curves after forecast.}
    \end{subfigure}
    \caption{\textbf{Shared Structure in Forecast Residuals.}
    Visualization on ETTh1 with iTransformer as the source forecaster.
    While raw channels are heterogeneous, forecast residuals exhibit structured patterns across channels.
    This suggests that independent channel-wise correction may be redundant.
    }
    \label{fig:intro}
\end{figure}

A key design choice in TSF-TTA is how to model test-time adaptation signals.
Since these signals are given by forecast residuals, i.e., the differences between model predictions and the corresponding ground truth, their channel-space structure directly affects adapter design.
In practice, most existing methods model these signals with channel-wise adapters, generating independent corrections for each channel~\citep{kim2025battling, medeiros2025accurate, grover2025shiftaware, im2026cosa}.
This design implicitly assumes that forecast residuals are inherently channel-specific and thus require independent modeling.
However, such independent parameterization can be redundant in TSF-TTA: the calibration modules are updated online from a limited number of sequentially revealed observations, and channel-wise adapters spend this limited supervision on repeatedly fitting correction patterns that are shared across channels.
Figure~\ref{fig:intro} shows that although raw channel behaviors are heterogeneous, their forecast residuals reveal common components across channels.
Since TSF-TTA aims to correct these residuals, the shared structure observed in the residuals suggests that the required corrections need not be fully channel-specific.
This raises the following research question: \emph{Are residual corrections in TSF-TTA truly channel-specific, or do they exhibit a shared structure that can be modeled more efficiently?}

In this work, we show that forecast residuals often exhibit structured patterns across channels, while channel-specific variations are relatively limited.
This indicates that fully independent channel-wise adapters are often over-parameterized for modeling corrections under limited online supervision.
Motivated by this insight, we propose the \textbf{D}ual-\textbf{B}ranch \textbf{C}alibration \textbf{M}odule (DBCM), which reformulates the correction in TSF-TTA by explicitly decomposing adaptation into shared and channel-specific components.
DBCM consists of a shared adapter that captures common correction patterns and a channel-wise low-rank adapter that models residual variations.
By constraining the channel-wise branch with low-rank parameterization, our design reduces redundant capacity while preserving flexibility for channel-specific corrections.
We further incorporate channel mixing for cross-channel context and nonlinear gated calibration for input-dependent modulation, and validate the effectiveness of the resulting module across diverse settings.

Our contributions are summarized as follows:
\begin{itemize}[leftmargin=10pt]
    \item We provide a new empirical perspective on TSF-TTA by analyzing forecast residuals in the channel-space. Our analysis shows that forecast residuals often exhibit substantial shared structure in the channel space, suggesting that fully independent channel-wise parameterization may be redundant.

    \item We propose the \textbf{D}ual-\textbf{B}ranch \textbf{C}alibration \textbf{M}odule (DBCM), which combines adaptation capacity shared across channels with limited, low-rank channel-specific flexibility. This design reduces redundant channel-wise capacity while preserving flexibility, and is further enhanced by channel mixing and nonlinear gated calibration for input-dependent modulation.

    \item We conduct extensive experiments across diverse datasets, forecasting horizons, and backbone architectures, demonstrating that DBCM consistently outperforms existing TSF-TTA methods while remaining relatively parameter-efficient.
\end{itemize}

\section{Related Work}

\paragraph{Test-Time Adaptation for Time Series Forecasting.}
TSF-TTA has emerged as an effective paradigm for handling non-stationarity.
TAFAS~\citep{kim2025battling} adapts a pre-trained model during inference using partially observed ground truth with a gated calibration module, while keeping the source forecaster fixed.
PETSA~\citep{medeiros2025accurate} improves efficiency by introducing low-rank adapters to calibrate input and output with dynamic gating.
DynaTTA~\citep{grover2025shiftaware} proposes a dynamic adaptation framework that estimates distribution shifts in real time and adjusts the adaptation rate accordingly.
COSA~\citep{im2026cosa} introduces a context-aware output-space adapter that directly corrects forecast residuals.
While the existing methods rely on channel-wise calibration,
we observe that forecast residuals often exhibit a shared low-dimensional structure across channels, indicating that such designs may be over-parameterized.
Building on this, we propose DBCM, which decomposes adaptation into shared and channel-specific components, reducing redundant channel-wise capacity while preserving flexibility through a low-rank formulation.
Additional related work is discussed in Appendix~\ref{app:related_work}.

\section{Method}
In this section, we present the Dual-Branch Calibration Module (DBCM), which, unlike the fully channel-wise adaptation of TAFAS and PETSA, combines a shared correction branch with a channel-wise low-rank branch.

\subsection{Problem Setting}
We consider a multivariate time series \(\{\mathbf{x}_t\}_{t=1}^{T}\), where \(\mathbf{x}_t \in \mathbb{R}^{C}\) denotes the \(C\)-channel observation at time \(t\).
Given a lookback window size \(L\) and a forecasting horizon \(H\), we define the input and target as
\(\mathbf{X}_t = \mathbf{x}_{t-L+1:t} \in \mathbb{R}^{L \times C}\) and
\(\mathbf{Y}_t = \mathbf{x}_{t+1:t+H} \in \mathbb{R}^{H \times C}\), respectively.
A source forecaster \(f_\theta\) produces predictions
\(\hat{\mathbf{Y}}_t = f_\theta(\mathbf{X}_t)\). At test time, instances arrive sequentially under possible distribution shifts.
Following prior methods~\citep{kim2025battling, medeiros2025accurate}, \(\mathbf{Y}_t\) is unavailable at prediction time but becomes partially observable as future time steps are revealed, and all methods are evaluated under the same protocol.
Following TAFAS~\citep{kim2025battling}, the reported MSE is computed on the adjusted predictions, identically for all methods.
In TSF-TTA, only calibration modules are adapted using the revealed test stream, while the source forecaster \(f_\theta\) remains frozen to preserve pre-trained knowledge.

\begin{figure}[t]
    \centering
    \includegraphics[width=1.0\columnwidth]{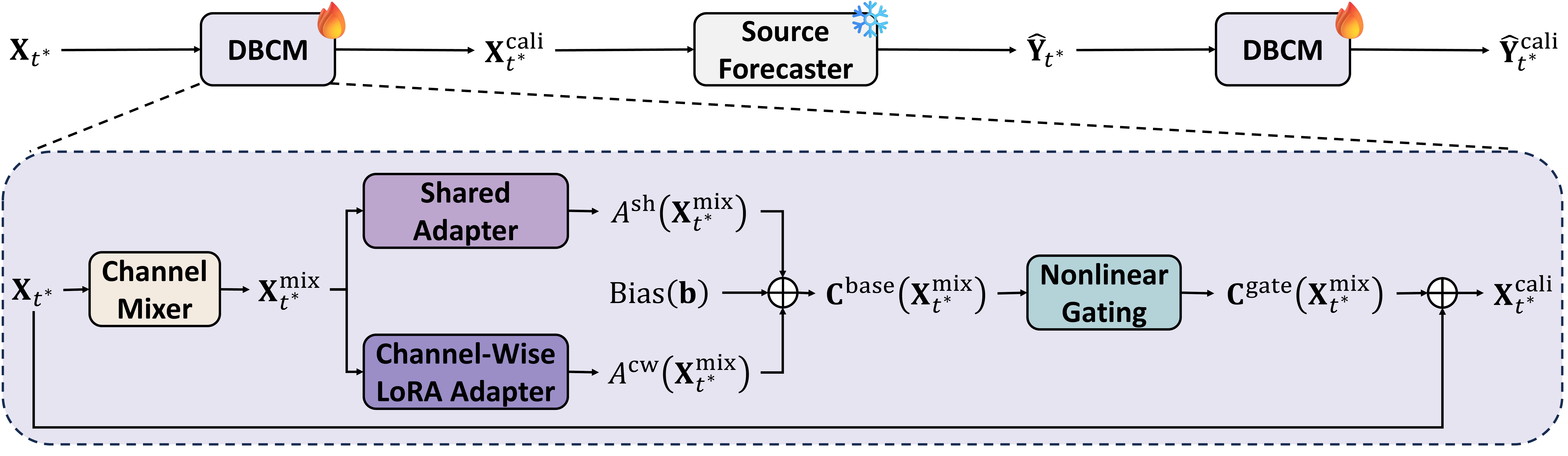}
    \vspace{-5pt}
    \caption{\textbf{Overview of the Dual-Branch Calibration Module (DBCM).}
Given an input sequence, DBCM first applies channel mixing and processes the mixed sequence with a shared adapter and a channel-wise LoRA adapter.
Their outputs are combined with a bias term, modulated by nonlinear gating, and added through a residual path to produce the calibrated sequence.
The same DBCM structure is used on the output-side to calibrate the forecast.}
    \vspace{-8pt}
    \label{fig:sr_tta_architecture}
\end{figure}

\subsection{Dual-Branch Calibration Module}

\paragraph{Overall Framework.}
To address distribution shift at test time, we adapt a pre-trained forecaster
by applying calibration modules to both the input and output sides.
Given an input sequence, the input-side DBCM first calibrates the input, which is then passed through the frozen source forecaster to produce a prediction.
The output-side DBCM subsequently calibrates the prediction.
Here, the input-side DBCM maps distribution-shifted test inputs closer to the input distribution that the pre-trained forecaster was trained to handle, while the output-side DBCM adjusts the forecaster's prediction toward the continuously evolving test distribution~\citep{kim2025battling}.
Formally, the calibration process is defined as
\begin{equation}
\mathbf{X}_{t^*}^{\mathrm{cali}} = C^{\mathrm{in}}(\mathbf{X}_{t^*}), \quad
\hat{\mathbf{Y}}_{t^*} = f_\theta(\mathbf{X}_{t^*}^{\mathrm{cali}}), \quad
\hat{\mathbf{Y}}_{t^*}^{\mathrm{cali}} = C^{\mathrm{out}}(\hat{\mathbf{Y}}_{t^*}),
\end{equation}
where \(C^{\mathrm{in}}\) and \(C^{\mathrm{out}}\) denote the input-side and output-side DBCMs, respectively, and \(t^*\) denotes the time step at which the update occurs.
The overall pipeline is illustrated in Figure~\ref{fig:sr_tta_architecture}.

We describe DBCM on the input side, as the output-side module has the same architecture, with the temporal dimension adjusted from \(L\) to \(H\).
The module consists of a linear channel mixer followed by two correction branches---a shared adapter and a channel-wise low-rank adapter---and a nonlinear gating mechanism.

\paragraph{Linear Channel Mixer.} To capture cross-channel interactions~\citep{chen2023tsmixer} before the subsequent calibration steps, a channel mixing transformation is first applied:
\begin{equation}
\mathbf{X}^{\text{mix}}_{t^*} = \mathbf{X}_{t^*} + \mathbf{X}_{t^*}\mathbf{W}^{\text{mix}},
\end{equation}
where \(\mathbf{W}^{\text{mix}} \in \mathbb{R}^{C \times C}\) is a learnable parameter.
The mixed representation is then passed to the shared and channel-wise branches, providing both branches with cross-channel context.
In effect, the channel mixer helps reveal shared latent structure in the channel-space, allowing the subsequent adapters to better exploit common correction patterns.
This interpretation is consistent with our isolated analysis in Appendix~\ref{app:channel_mixer}, where the channel mixer makes forecast residuals more concentrated in a low-rank channel subspace while reducing their overall magnitude.

\paragraph{Shared Adapter.} As illustrated in Figure~\ref{fig:intro} and analyzed in Section~\ref{sec:analysis}, forecast residuals exhibit shared structure across channels, suggesting that fully independent adaptation parameters for every channel may be redundant.
We therefore share adaptation parameters across channels by applying a single temporal transformation to all channels:
\begin{equation}
A^{\text{sh}}(\mathbf{X}^{\text{mix}}_{t^*}) = \mathbf{W}^{\text{sh}} \mathbf{X}^{\text{mix}}_{t^*},
\end{equation}
where \(\mathbf{W}^{\text{sh}} \in \mathbb{R}^{L \times L}\) is a learnable parameter applied uniformly to all channels.
Intuitively, when channels are driven by a few common components, a shared temporal transformation corrects these components jointly across channels, whereas fully channel-wise adapters must learn them separately for each channel.
By applying the same temporal transformation to every channel, the shared adapter is designed to capture common correction patterns with substantially fewer parameters than fully channel-wise adapters, and its effectiveness is empirically validated in Section~\ref{sec:ablation}.

\paragraph{Channel-Wise LoRA Adapter.}
Although the shared adapter captures common correction patterns across channels, channel-specific residual variations may remain.
To model them without using a fully parameterized channel-wise adapter, we introduce a channel-wise low-rank branch.
For each channel \(c\), an independent low-rank temporal transformation is applied to the channel-mixed sequence:
\begin{equation}
A^{\text{cw}}(\mathbf{X}^{\text{mix}}_{t^*}) = \operatorname{Concat}\!\left(
\left\{(\mathbf{U}_c \mathbf{V}_c)\,\mathbf{x}^{\text{mix}}_{{t^*},c}\right\}_{c=1}^{C}
\right),
\end{equation}
where \(\mathbf{x}^{\text{mix}}_{{t^*}, c} \in \mathbb{R}^{L}\) denotes the \(c\)-th channel of \(\mathbf{X}^{\text{mix}}_{t^*}\), and \(\mathbf{U}_c \in \mathbb{R}^{L \times r}\) and \(\mathbf{V}_c \in \mathbb{R}^{r \times L}\) define a channel-specific low-rank parameterization with rank \(r \ll L\). \(\operatorname{Concat}(\cdot)\) concatenates the corrections along the channel dimension.
We constrain this branch to be low rank, limiting its expressive capacity to complement the shared adapter while preserving flexibility for channel-specific corrections.

\paragraph{Nonlinear Gated Calibration.} The \emph{base correction} is computed by combining the outputs from the shared and channel-wise adapters:
\begin{equation}
\mathbf{C}^{\mathrm{base}}(\mathbf{X}^{\mathrm{mix}}_{t^*})
=
A^{\mathrm{sh}}(\mathbf{X}^{\mathrm{mix}}_{t^*})
+
A^{\mathrm{cw}}(\mathbf{X}^{\mathrm{mix}}_{t^*})
+
\mathbf{b},
\end{equation}
where \(\mathbf{b} \in \mathbb{R}^{C}\) is a learnable channel-wise bias parameter, which is broadcast along the temporal dimension to \(\mathbb{R}^{L \times C}\).

While the base correction captures linear correction patterns, we introduce an input-dependent nonlinear gate to model variations that cannot be captured by a purely linear correction:
\begin{equation}
\mathbf{G}(\mathbf{X}^{\mathrm{mix}}_{t^*})
=
\sigma\!\left(\mathbf{W}^{\mathrm{gate}} \mathbf{X}^{\mathrm{mix}}_{t^*} + \mathbf{b}^{\mathrm{gate}}\right),
\end{equation}
where \(\sigma(\cdot)\) denotes the sigmoid function,
\(\mathbf{W}^{\mathrm{gate}} \in \mathbb{R}^{L \times L}\) and
\(\mathbf{b}^{\mathrm{gate}} \in \mathbb{R}^{L}\) are learnable parameters, with
\(\mathbf{b}^{\mathrm{gate}}\) broadcast to \(\mathbb{R}^{L \times C}\) along the channel dimension. Using this gate, we construct the \emph{gated correction}
\begin{equation}
\mathbf{C}^{\mathrm{gate}}(\mathbf{X}^{\mathrm{mix}}_{t^*})
=
\mathbf{C}^{\mathrm{base}}(\mathbf{X}^{\mathrm{mix}}_{t^*})
+
\mathbf{C}^{\mathrm{base}}(\mathbf{X}^{\mathrm{mix}}_{t^*})
\odot
\mathbf{G}(\mathbf{X}^{\mathrm{mix}}_{t^*}),
\end{equation}
where \(\odot\) denotes a Hadamard product.

The calibrated input is then given by
\begin{equation}
\mathbf{X}_{t^*}^{\mathrm{cali}}
=
\mathbf{X}_{t^*}
+
\tanh(\mathbf{g})
\odot
\mathbf{C}^{\mathrm{gate}}(\mathbf{X}^{\mathrm{mix}}_{t^*}),
\end{equation}
where \(\mathbf{g} \in \mathbb{R}^{C}\) is a learnable channel-wise gating parameter, and
\(\tanh(\mathbf{g})\) is broadcast to \(\mathbb{R}^{L \times C}\) along the temporal dimension.
Our two-level gating design separates fine-grained correction modulation from coarse channel-wise scaling used in prior TSF-TTA methods.
Following prior TSF-TTA approaches~\citep{kim2025battling, medeiros2025accurate, grover2025shiftaware, im2026cosa}, the outer gate \(\tanh(\mathbf{g})\) controls the overall channel-wise contribution of the correction.
PETSA~\citep{medeiros2025accurate} further demonstrates that input-dependent dynamic gating is effective for adapting calibration strength under test-time distribution shifts.
However, such gating primarily acts as a scaling mechanism for the calibrated correction.
In contrast, our inner nonlinear gate modulates the correction values themselves, enabling input-dependent adjustment of the temporal correction pattern before the outer channel-wise gate is applied.
Following the intuition of gated linear units~\citep{dauphin2017language}, this formulation preserves the linear correction path through \(\mathbf{C}^{\mathrm{base}}\), while
\(\mathbf{C}^{\mathrm{base}}\odot\mathbf{G}\) introduces input-dependent nonlinear modulation through the sigmoid gate.

The output-side module follows the same design, with the temporal dimension adjusted from the input length \(L\) to the prediction horizon \(H\).
The complete TSF-TTA algorithm is provided in Appendix~\ref{app:algorithm}.

\section{Analysis}
\label{sec:analysis}

In this section, we analyze TSF-TTA signals in the channel-space to provide empirical evidence for the design of DBCM.
Specifically, we characterize whether forecast residuals, which serve as the supervision signal for test-time adaptation, are concentrated along shared channel directions, and examine whether adapter corrections align with these residual patterns.
Additional analyses, including the effects of the channel-wise LoRA adapter, are provided in Appendix~\ref{app:additional_analysis}.

\paragraph{Channel-Space Analysis Protocol.}
For each forecasting window \(\mathbf{X}_t \in \mathbb{R}^{L \times C}\),
we consider a channel-wise signal associated with the window, such as a forecast residual or an adapter correction.
The forecast residual is \(\mathbf{R}_t = \mathbf{Y}_t - f_\theta(\mathbf{X}_t)\), and the adapter correction is defined as the change in the prediction induced by the calibration modules,
\begin{equation}
\boldsymbol{\Delta}_t = \hat{\mathbf{Y}}^{\mathrm{cali}}_t - f_\theta(\mathbf{X}_t) = C^{\mathrm{out}}\big(f_\theta(C^{\mathrm{in}}(\mathbf{X}_t))\big) - f_\theta(\mathbf{X}_t),
\end{equation}
which includes both the effect of the input-side module propagated through the frozen forecaster and the correction directly applied by the output-side module.
We stack these signals over the \(B\) evaluated forecasting instances and reshape them into a channel-space matrix \(\mathbf{S} \in \mathbb{R}^{C \times (BH)}\), where each row corresponds to a channel.
We then analyze the channel-space spectrum using the centered Gram matrix \(\boldsymbol{\Sigma}=\tilde{\mathbf{S}}\tilde{\mathbf{S}}^{\top}\), where \(\tilde{\mathbf{S}}\) is obtained by subtracting the mean of each row (channel) of \(\mathbf{S}\).
Let \(\lambda_1 \ge \lambda_2 \ge \cdots \ge \lambda_C\) denote the eigenvalues of \(\boldsymbol{\Sigma}\).
The top-\(k\) explained variance ratio is defined as
\begin{equation}
\mathrm{EVR}@k = \frac{\sum_{i=1}^{k}\lambda_i}{\sum_{i=1}^{C}\lambda_i},
\end{equation}
and \(\mathrm{Rank}@90\) is the smallest \(k\) such that \(\mathrm{EVR}@k \ge 0.9\).
Since the number of channels differs across datasets, we report channel-normalized metrics: \(\mathrm{EVR}@30\%\) denotes \(\mathrm{EVR}@k\) with \(k=\lceil 0.3C \rceil\), and \(\mathrm{Rank}@90/C\) normalizes \(\mathrm{Rank}@90\) by the number of channels, both reported as percentages.
A high \(\mathrm{EVR}@30\%\) and a small \(\mathrm{Rank}@90/C\) indicate that the signal is concentrated along a few shared channel directions.
To compare forecast residuals with adapter corrections, let \(\mathbf{U}^{R}, \mathbf{U}^{\Delta} \in \mathbb{R}^{C \times k}\) denote orthonormal bases of the top-\(k\) principal channel directions of the forecast residuals and adapter corrections, i.e., the leading eigenvectors of their centered Gram matrices.
We measure their subspace alignment by the projection similarity
\begin{equation}
\mathrm{Align}(\mathbf{U}^{R}, \mathbf{U}^{\Delta}) = \frac{1}{k}\,\big\|(\mathbf{U}^{R})^{\top}\mathbf{U}^{\Delta}\big\|_F^2 = \frac{1}{k}\,\mathrm{tr}\big(\mathbf{P}^{R}\mathbf{P}^{\Delta}\big),
\end{equation}
where \(\mathbf{P}^{R}\) and \(\mathbf{P}^{\Delta}\) are the orthogonal projections onto the two subspaces, and we use \(k=2\).
The score equals the mean squared cosine of the principal angles between the subspaces, is invariant to the sign and rotation of the basis vectors, and ranges from 0 to 1, with an expected value of \(k/C\) for randomly drawn subspaces.

\paragraph{Forecast Residual Structure.}
Since calibration modules are optimized against the forecast residual \(\mathbf{R}_t\), its channel-space spectrum indicates whether residual variation is concentrated in a few shared channel directions or dispersed across channels.
This distinction is relevant to adapter design: shared residual components need not be modeled separately for each channel, whereas residual variation not explained by those components motivates channel-specific flexibility.
As shown in Table~\ref{tab:residual_spectra_backbones}, the residual spectra are consistent across backbone architectures, with \(\mathrm{EVR}@30\%\) values around 93\% and \(\mathrm{Rank}@90/C\) values around 42\%.
These values are far from those of isotropic residuals evenly spread over all channel directions, which would yield an \(\mathrm{EVR}@30\%\) of 33--43\% and a \(\mathrm{Rank}@90/C\) of 90--100\%, suggesting that the dominant correction patterns are largely shared across channels.
Fully independent channel-wise adapters may redundantly model these shared residual patterns with separate parameters for each channel.
Rather than prescribing a specific architecture, this observation serves as empirical motivation for sharing adaptation capacity across channels, whose effectiveness we validate through the ablation in Section~\ref{sec:ablation}.

Table~\ref{tab:residual_spectra_datasets} further shows that the degree of concentration varies across datasets.
Weather exhibits the strongest concentration, with an \(\mathrm{EVR}@30\%\) of 98.2\% and a \(\mathrm{Rank}@90/C\) of 27.4\%, whereas Exchange Rate exhibits the weakest concentration, with an \(\mathrm{EVR}@30\%\) of 85.7\% and a \(\mathrm{Rank}@90/C\) of 50.0\%.
Exchange Rate still remains structured, but its lower \(\mathrm{EVR}@30\%\) and higher \(\mathrm{Rank}@90/C\) indicate that a non-negligible portion of residual variation lies outside the dominant shared directions.
In such cases, a purely shared adapter may be too restrictive, while a full channel-wise adapter introduces unnecessary capacity.
DBCM addresses this trade-off with two branches: a shared branch for patterns common across channels and a channel-wise LoRA branch for residual variations not captured by the shared branch.
The strength of such structure depends on both the data-generating process and the forecaster~\citep{west1996asymptotic}, and it can arise when many channels share a few factors that are incompletely modeled~\citep{zheng2024multivariate, salinas2019high}.

\begin{table*}[t]
\centering
\caption{
\textbf{Channel-Space Analysis.}
Channel-space concentration of forecast residuals across backbones and datasets, and the subspace alignment between forecast residuals and adapter corrections.
}
\label{tab:analysis_spectrum}

\begin{subtable}[t]{0.345\textwidth}
\centering
\caption{Residual spectra by backbone.}
\label{tab:residual_spectra_backbones}
\footnotesize
\setlength{\tabcolsep}{0pt}
\renewcommand{\arraystretch}{0.9}
\begin{tabular*}{\linewidth}{@{\extracolsep{\fill}}lcc@{}}
\toprule
\textbf{Backbone} & \scalebox{0.93}{\(\mathrm{EVR}@30\%\)} & \scalebox{0.93}{\(\mathrm{Rank}@90/C\)} \\
\midrule
DLinear      & 93.4 & 41.5 \\
FreTS        & 93.0 & 41.5 \\
MICN         & 93.2 & 41.5 \\
OLS          & 93.3 & 41.5 \\
PatchTST     & 92.9 & 41.5 \\
iTransformer & 92.8 & 42.1 \\
\midrule
Average      & 93.1 & 41.6 \\
\bottomrule
\end{tabular*}
\end{subtable}
\hfill
\begin{subtable}[t]{0.345\textwidth}
\centering
\caption{Residual spectra by dataset.}
\label{tab:residual_spectra_datasets}
\footnotesize
\setlength{\tabcolsep}{0pt}
\renewcommand{\arraystretch}{0.9}
\begin{tabular*}{\linewidth}{@{\extracolsep{\fill}}lcc@{}}
\toprule
\textbf{Dataset} & \scalebox{0.93}{\(\mathrm{EVR}@30\%\)} & \scalebox{0.93}{\(\mathrm{Rank}@90/C\)} \\
\midrule
ETTh1    & 94.7 & 42.9 \\
ETTh2    & 93.1 & 42.9 \\
ETTm1    & 95.0 & 42.9 \\
ETTm2    & 91.8 & 43.5 \\
Exchange & 85.7 & 50.0 \\
Weather  & 98.2 & 27.4 \\
\midrule
Average  & 93.1 & 41.6 \\
\bottomrule
\end{tabular*}
\end{subtable}
\hfill
\begin{subtable}[t]{0.25\textwidth}
\centering
\caption{Subspace alignment.}
\label{tab:subspace_alignment}
\footnotesize
\setlength{\tabcolsep}{0pt}
\renewcommand{\arraystretch}{0.95}
\begin{tabular*}{\linewidth}{@{\extracolsep{\fill}}lc@{}}
\toprule
\textbf{Adapter} & \textbf{Alignment} \\
\midrule
Oracle       & 1.000 \\
Random       & 0.246 \\
\midrule
TAFAS        & 0.671 \\
TAFAS-Shared & 0.723 \\
PETSA        & 0.666 \\
\midrule
DBCM (Ours)  & \textbf{0.774} \\
\bottomrule
\end{tabular*}
\end{subtable}

\vspace{-8pt}
\end{table*}

\paragraph{Residual--Correction Alignment.} We further analyze how adapter corrections obtained during TSF-TTA align with forecast residuals.
Table~\ref{tab:subspace_alignment} reports the alignment between their principal channel directions.
Oracle denotes the ideal adapter correction, defined as the forecast residual itself, and therefore achieves an alignment score of 1 by construction, while Random denotes the expected score of randomly drawn subspaces.
Across methods, adapter corrections consistently align with forecast residuals, yielding alignment scores far above the random baseline.
This suggests that the adapted corrections effectively reflect the residual structure.
Notably, TAFAS-Shared, a TAFAS variant that uses a single shared adapter instead of channel-wise adapters, achieves higher alignment than TAFAS (0.723 vs.\ 0.671).
This further supports the residual-spectrum analysis: the dominant shared component of the correction can be largely captured without fully independent channel-wise parameterization.
Among the compared methods, DBCM achieves the strongest alignment, indicating that its corrections are most closely aligned with the leading residual subspace.
This result is consistent with decomposing correction into a shared branch and a low-rank channel-wise branch.
Appendix~\ref{app:dataset_wise_analysis} provides additional dataset-wise alignment scores and analyses.

\section{Experiments}

We evaluate the effectiveness of DBCM under test-time distribution shifts on multiple benchmark datasets.
We consider diverse backbone architectures and forecasting horizons to verify the general applicability of our method.
Additional experimental results, including comparisons with DynaTTA~\citep{grover2025shiftaware} and COSA~\citep{im2026cosa}, are provided in Appendix~\ref{app:additional_exp}.

\subsection{Experimental Setup}
\begin{table}[!t]
\centering
\caption{\textbf{Prediction Performance Comparison Across Methods.} Test MSE is reported for six models on six datasets with four prediction horizons. Bold entries indicate the best performance among methods, while underlined entries denote the second-best. Avg. Imp. denotes the relative MSE reduction (\%) over the frozen baseline, averaged over all dataset--horizon settings. Full results with standard deviations are provided in Table~\ref{tab:main_std}.
}
\label{tab:main}
\setlength{\tabcolsep}{3pt}
\renewcommand{\arraystretch}{0.70}
\setlength{\aboverulesep}{1.5pt}
\setlength{\belowrulesep}{1.5pt}
\setlength{\abovetopsep}{1.5pt}
\setlength{\belowbottomsep}{1.5pt}
\setlength{\cmidrulesep}{1.5pt}
\scriptsize
{
\resizebox{\textwidth}{!}{
\begin{tabular}{c c c c c c c c c c c c c c}
\toprule
\multicolumn{2}{c}{} & \multicolumn{4}{c}{\textbf{iTransformer}} & \multicolumn{4}{c}{\textbf{DLinear}} & \multicolumn{4}{c}{\textbf{FreTS}} \\
\cmidrule(lr){3-6}\cmidrule(lr){7-10}\cmidrule(lr){11-14}
\multicolumn{2}{c}{} & Baseline & TAFAS & PETSA & DBCM & Baseline & TAFAS & PETSA & DBCM & Baseline & TAFAS & PETSA & DBCM \\
\midrule
\multirow{4}{*}{\rotatebox[origin=c]{90}{\textbf{ETTh1}}} & 96 & .4349 & .4308 & \underline{.4284} & \textbf{.4228} & .4510 & .4425 & \underline{.4400} & \textbf{.4318} & .4360 & \underline{.4295} & .4296 & \textbf{.4242} \\
 & 192 & .4937 & .4894 & \underline{.4886} & \textbf{.4805} & .5042 & .4946 & \underline{.4943} & \textbf{.4835} & .4886 & \underline{.4819} & .4842 & \textbf{.4764} \\
 & 336 & .5406 & \underline{.5382} & .5392 & \textbf{.5322} & .5508 & \underline{.5454} & .5455 & \textbf{.5373} & .5466 & \underline{.5394} & .5462 & \textbf{.5354} \\
 & 720 & .6914 & \underline{.6520} & .6535 & \textbf{.6298} & .7019 & .6713 & \underline{.6617} & \textbf{.6487} & .7145 & .6839 & \underline{.6795} & \textbf{.6565} \\ \midrule
\multirow{4}{*}{\rotatebox[origin=c]{90}{\textbf{ETTh2}}} & 96 & .2430 & .2406 & \underline{.2403} & \textbf{.2388} & .2294 & \underline{.2274} & .2280 & \textbf{.2268} & .2376 & \textbf{.2347} & \underline{.2359} & \textbf{.2347} \\
 & 192 & .2973 & .2931 & \underline{.2918} & \textbf{.2914} & .2835 & \textbf{.2793} & .2840 & \underline{.2794} & .2838 & \underline{.2776} & .2805 & \textbf{.2761} \\
 & 336 & .3448 & \underline{.3328} & .3334 & \textbf{.3301} & .3250 & \underline{.3174} & .3185 & \textbf{.3152} & .3184 & \underline{.3051} & .3117 & \textbf{.3048} \\
 & 720 & .4258 & .3999 & \underline{.3987} & \textbf{.3939} & .4149 & .3913 & \underline{.3878} & \textbf{.3820} & .3951 & \underline{.3690} & .3812 & \textbf{.3653} \\ \midrule
\multirow{4}{*}{\rotatebox[origin=c]{90}{\textbf{ETTm1}}} & 96 & .3763 & .3606 & \underline{.3601} & \textbf{.3415} & .3705 & \underline{.3496} & .3524 & \textbf{.3420} & .3618 & \underline{.3524} & .3525 & \textbf{.3420} \\
 & 192 & .4371 & .4184 & \underline{.4145} & \textbf{.4038} & .4420 & \underline{.4159} & .4175 & \textbf{.4056} & .4304 & .4177 & \underline{.4168} & \textbf{.4060} \\
 & 336 & .5029 & .4778 & \underline{.4739} & \textbf{.4628} & .5172 & .4796 & \underline{.4795} & \textbf{.4626} & .4928 & .4752 & \underline{.4719} & \textbf{.4618} \\
 & 720 & .5713 & .5484 & \underline{.5472} & \textbf{.5276} & .5927 & \underline{.5482} & .5526 & \textbf{.5321} & .5609 & \underline{.5399} & .5404 & \textbf{.5274} \\ \midrule
\multirow{4}{*}{\rotatebox[origin=c]{90}{\textbf{ETTm2}}} & 96 & .1555 & \underline{.1542} & .1543 & \textbf{.1536} & .1589 & \underline{.1577} & .1579 & \textbf{.1573} & .1563 & \underline{.1547} & .1549 & \textbf{.1536} \\
 & 192 & .1938 & \underline{.1926} & .1928 & \textbf{.1923} & .1929 & \underline{.1913} & \underline{.1913} & \textbf{.1907} & .1914 & \underline{.1896} & .1898 & \textbf{.1880} \\
 & 336 & .2345 & .2441 & \underline{.2325} & \textbf{.2312} & .2325 & \underline{.2290} & .2293 & \textbf{.2289} & .2310 & \underline{.2275} & .2305 & \textbf{.2259} \\
 & 720 & .3101 & \underline{.3012} & .3033 & \textbf{.2990} & .3061 & .2967 & \underline{.2962} & \textbf{.2956} & .3026 & \textbf{.2924} & .3003 & \underline{.2930} \\ \midrule
\multirow{4}{*}{\rotatebox[origin=c]{90}{\parbox{0.96cm}{\centering\textbf{Exchange}}}} & 96 & .0916 & .0883 & \underline{.0879} & \textbf{.0858} & .0811 & \textbf{.0789} & \underline{.0793} & \textbf{.0789} & .0827 & \textbf{.0784} & .0815 & \underline{.0794} \\
 & 192 & .1914 & \underline{.1834} & .1867 & \textbf{.1770} & .1708 & .1643 & \underline{.1641} & \textbf{.1626} & .1787 & \underline{.1667} & .1765 & \textbf{.1659} \\
 & 336 & .3375 & \underline{.3146} & .3243 & \textbf{.2844} & .3205 & \underline{.2883} & .2931 & \textbf{.2642} & .3265 & \underline{.2922} & .3198 & \textbf{.2659} \\
 & 720 & .9433 & \underline{.8388} & .8894 & \textbf{.8182} & .8383 & .8309 & \underline{.8109} & \textbf{.7066} & .8251 & \underline{.7606} & .8141 & \textbf{.6700} \\ \midrule
\multirow{4}{*}{\rotatebox[origin=c]{90}{\textbf{Weather}}} & 96 & .1719 & \underline{.1642} & .1661 & \textbf{.1630} & .1936 & .1788 & \underline{.1769} & \textbf{.1732} & .1809 & \underline{.1693} & .1713 & \textbf{.1677} \\
 & 192 & .2233 & \underline{.2104} & .2129 & \textbf{.2096} & .2383 & \underline{.2236} & .2255 & \textbf{.2194} & .2266 & \underline{.2109} & .2179 & \textbf{.2101} \\
 & 336 & .2800 & \underline{.2610} & .2646 & \textbf{.2608} & .2913 & \underline{.2714} & .2755 & \textbf{.2697} & .2806 & \underline{.2609} & .2675 & \textbf{.2602} \\
 & 720 & .3576 & \underline{.3404} & .3498 & \textbf{.3360} & .3643 & \underline{.3490} & .3501 & \textbf{.3391} & .3557 & \underline{.3364} & .3536 & \textbf{.3338} \\ \midrule
\multicolumn{2}{c}{\textbf{Average}} & .3687 & \underline{.3531} & .3556 & \textbf{.3444} & .3655 & .3509 & \underline{.3505} & \textbf{.3389} & .3585 & \underline{.3436} & .3503 & \textbf{.3343} \\
\arrayrulecolor{black!35}\midrule\arrayrulecolor{black}
\multicolumn{2}{c}{\textbf{Avg. Rank}} & 3.96 & \underline{2.50} & 2.54 & \textbf{1.00} & 3.96 & \underline{2.33} & 2.65 & \textbf{1.06} & 4.00 & \underline{2.02} & 2.88 & \textbf{1.10} \\
\arrayrulecolor{black!35}\midrule\arrayrulecolor{black}
\multicolumn{2}{c}{\textbf{Avg. Imp.}} & -- & \underline{3.55} & 3.19 & \textbf{5.75} & -- & \underline{3.93} & 3.83 & \textbf{6.22} & -- & \underline{4.10} & 2.15 & \textbf{5.83} \\
\bottomrule
\end{tabular}
}
\vspace{0.8em}
\resizebox{\textwidth}{!}{
\begin{tabular}{c c c c c c c c c c c c c c}
\toprule
\multicolumn{2}{c}{} & \multicolumn{4}{c}{\textbf{PatchTST}} & \multicolumn{4}{c}{\textbf{OLS}} & \multicolumn{4}{c}{\textbf{MICN}} \\
\cmidrule(lr){3-6}\cmidrule(lr){7-10}\cmidrule(lr){11-14}
\multicolumn{2}{c}{} & Baseline & TAFAS & PETSA & DBCM & Baseline & TAFAS & PETSA & DBCM & Baseline & TAFAS & PETSA & DBCM \\
\midrule
\multirow{4}{*}{\rotatebox[origin=c]{90}{\textbf{ETTh1}}} & 96 & .4294 & \underline{.4232} & .4255 & \textbf{.4194} & .4511 & \underline{.4420} & .4436 & \textbf{.4321} & .4573 & .4471 & \underline{.4453} & \textbf{.4342} \\
 & 192 & .4904 & \underline{.4805} & .4849 & \textbf{.4757} & .5044 & \underline{.4923} & .5040 & \textbf{.4836} & .5145 & \underline{.4996} & .5063 & \textbf{.4900} \\
 & 336 & .5605 & \underline{.5490} & .5600 & \textbf{.5410} & .5510 & \underline{.5415} & .5505 & \textbf{.5383} & .5715 & \underline{.5596} & .5708 & \textbf{.5553} \\
 & 720 & .6904 & \underline{.6643} & .6878 & \textbf{.6417} & .6996 & \underline{.6654} & .6728 & \textbf{.6514} & .7224 & \underline{.6916} & .7201 & \textbf{.6743} \\ \midrule
\multirow{4}{*}{\rotatebox[origin=c]{90}{\textbf{ETTh2}}} & 96 & .2309 & \underline{.2298} & .2308 & \textbf{.2291} & .2305 & \underline{.2283} & .2286 & \textbf{.2279} & .2346 & \underline{.2326} & .2333 & \textbf{.2321} \\
 & 192 & .2871 & \underline{.2802} & .2867 & \textbf{.2785} & .2838 & \underline{.2817} & .2822 & \textbf{.2802} & .2934 & \underline{.2868} & .2898 & \textbf{.2848} \\
 & 336 & .3311 & \underline{.3163} & .3303 & \textbf{.3157} & .3257 & \underline{.3186} & .3251 & \textbf{.3157} & .3380 & \underline{.3254} & .3309 & \textbf{.3250} \\
 & 720 & .4067 & \underline{.3886} & .3958 & \textbf{.3769} & .4164 & .3902 & \underline{.3883} & \textbf{.3831} & .4328 & \underline{.4126} & .4191 & \textbf{.4090} \\ \midrule
\multirow{4}{*}{\rotatebox[origin=c]{90}{\textbf{ETTm1}}} & 96 & .3869 & \underline{.3723} & .3829 & \textbf{.3469} & .3708 & \underline{.3520} & .3574 & \textbf{.3420} & .4063 & \underline{.3723} & .3756 & \textbf{.3576} \\
 & 192 & .4387 & \underline{.4220} & .4328 & \textbf{.4057} & .4437 & \underline{.4165} & .4230 & \textbf{.4054} & .4479 & .4315 & \underline{.4308} & \textbf{.4155} \\
 & 336 & .4957 & \underline{.4780} & .4939 & \textbf{.4593} & .5176 & \underline{.4786} & .4853 & \textbf{.4623} & .5154 & \underline{.4885} & .4889 & \textbf{.4676} \\
 & 720 & .5666 & \underline{.5390} & .5615 & \textbf{.5255} & .5918 & \underline{.5478} & .5519 & \textbf{.5317} & .5749 & \underline{.5404} & .5689 & \textbf{.5206} \\ \midrule
\multirow{4}{*}{\rotatebox[origin=c]{90}{\textbf{ETTm2}}} & 96 & .1554 & \underline{.1542} & .1552 & \textbf{.1534} & .1602 & .1593 & \underline{.1592} & \textbf{.1585} & .1592 & \underline{.1580} & .1582 & \textbf{.1574} \\
 & 192 & .1937 & \underline{.1924} & .1935 & \textbf{.1919} & .1935 & \underline{.1921} & \textbf{.1919} & \textbf{.1919} & .1955 & .1927 & \underline{.1923} & \textbf{.1914} \\
 & 336 & .2321 & \underline{.2292} & .2317 & \textbf{.2282} & .2331 & \textbf{.2298} & \underline{.2301} & \textbf{.2298} & .2421 & .2364 & \textbf{.2355} & \underline{.2357} \\
 & 720 & .3033 & \textbf{.2924} & .3005 & \underline{.2933} & .3066 & .2982 & \underline{.2977} & \textbf{.2964} & .3176 & \underline{.3040} & .3049 & \textbf{.3023} \\ \midrule
\multirow{4}{*}{\rotatebox[origin=c]{90}{\parbox{0.96cm}{\centering\textbf{Exchange}}}} & 96 & .0846 & \underline{.0811} & .0839 & \textbf{.0806} & .0814 & \underline{.0795} & .0808 & \textbf{.0793} & .0829 & \underline{.0804} & .0818 & \textbf{.0800} \\
 & 192 & .1805 & \underline{.1697} & .1787 & \textbf{.1683} & .1723 & \underline{.1640} & .1700 & \textbf{.1631} & .1886 & \underline{.1771} & .1853 & \textbf{.1742} \\
 & 336 & .5375 & \underline{.4585} & .4755 & \textbf{.4087} & .3226 & \underline{.2851} & .2876 & \textbf{.2613} & .3529 & \underline{.3116} & .3315 & \textbf{.2829} \\
 & 720 & .8814 & \underline{.8812} & .8814 & \textbf{.6609} & .8360 & \underline{.6485} & .8114 & \textbf{.5439} & 1.4040 & \textbf{1.1221} & 1.3355 & \underline{1.2209} \\ \midrule
\multirow{4}{*}{\rotatebox[origin=c]{90}{\textbf{Weather}}} & 96 & .1738 & \underline{.1672} & .1730 & \textbf{.1657} & .1955 & \underline{.1808} & .1819 & \textbf{.1763} & .1720 & \underline{.1699} & .1726 & \textbf{.1670} \\
 & 192 & .2207 & \underline{.2103} & .2183 & \textbf{.2073} & .2403 & \underline{.2219} & .2272 & \textbf{.2196} & .2222 & \underline{.2155} & .2237 & \textbf{.2148} \\
 & 336 & .2760 & \underline{.2598} & .2721 & \textbf{.2586} & .2920 & \underline{.2706} & .2772 & \textbf{.2697} & .2747 & \underline{.2626} & .2701 & \textbf{.2624} \\
 & 720 & .3512 & .3644 & \underline{.3410} & \textbf{.3315} & .3646 & \underline{.3443} & .3626 & \textbf{.3398} & .3491 & \underline{.3371} & .3546 & \textbf{.3298} \\ \midrule
\multicolumn{2}{c}{\textbf{Average}} & .3710 & \underline{.3585} & .3657 & \textbf{.3402} & .3660 & \underline{.3429} & .3538 & \textbf{.3326} & .3946 & \underline{.3690} & .3844 & \textbf{.3660} \\
\arrayrulecolor{black!35}\midrule\arrayrulecolor{black}
\multicolumn{2}{c}{\textbf{Avg. Rank}} & 3.94 & \underline{2.04} & 2.98 & \textbf{1.04} & 4.00 & \underline{2.15} & 2.81 & \textbf{1.04} & 3.88 & \underline{2.12} & 2.92 & \textbf{1.08} \\
\arrayrulecolor{black!35}\midrule\arrayrulecolor{black}
\multicolumn{2}{c}{\textbf{Avg. Imp.}} & -- & \underline{3.29} & 1.24 & \textbf{6.43} & -- & \underline{5.09} & 3.12 & \textbf{7.07} & -- & \underline{4.50} & 2.14 & \textbf{5.90} \\
\bottomrule
\end{tabular}
}
}
\end{table}

\paragraph{Datasets.} We conduct experiments on six standard multivariate time series forecasting benchmarks: ETTh1, ETTh2, ETTm1, ETTm2, Exchange Rate, and Weather. These datasets are widely used in prior long-term forecasting studies and exhibit diverse temporal dynamics~\citep{zhou2021informer,wu2021autoformer}, making them suitable for evaluating adaptation performance under distribution shift. We further consider Electricity in an additional experiment to evaluate scalability.
Given its substantially larger number of channels, it serves as a challenging testbed for assessing whether our method remains robust in high-dimensional multivariate settings. Detailed dataset information is provided in Appendix~\ref{app:data_description}.

\paragraph{Backbone Models.} We evaluate our method on a diverse set of backbone models.
These models cover a wide range of architectural paradigms for time series forecasting, including linear models (DLinear~\citep{zeng2022arete}, OLS~\citep{toner2024analysis}),
transformer-based models (iTransformer~\citep{liu2023itransformer}, PatchTST~\citep{nie2022time}), frequency-based methods (FreTS~\citep{yi2023frequency}), and convolutional architectures (MICN~\citep{wang2023micn}).
This diversity allows us to assess the general applicability of our approach across different modeling assumptions. In addition, we further validate our method on recent TSFMs, including Moirai~\citep{woo2024moirai}, Chronos~\citep{ansari2024chronos}, and Chronos-2~\citep{ansari2025chronos2}. These models are pre-trained on large-scale datasets and represent a different regime from standard supervised forecasting models.
Evaluating on these models demonstrates that our approach remains effective even in large-scale pre-trained settings.

\paragraph{Implementation Details.} Following TimesNet~\citep{wu2023timesnet}, we use a fixed lookback window $L = 96$ and prediction horizons $H \in \{96, 192, 336, 720\}$.
All datasets are split chronologically, with train/validation/test ratios of 0.6/0.2/0.2 for the ETT datasets and 0.7/0.1/0.2 for Exchange Rate, Weather, and Electricity.
We report mean squared error (MSE) as the primary evaluation metric.
In all experiments, we set the rank of the channel-wise LoRA adapter to \(r=8\) unless otherwise specified.
In the ablation studies, we use iTransformer~\citep{liu2023itransformer} as the backbone model.
To ensure robustness, results are averaged over three runs with different random seeds.

\subsection{Main Results}
We compare our method with state-of-the-art TSF-TTA approaches, TAFAS~\citep{kim2025battling} and PETSA~\citep{medeiros2025accurate}.
As shown in Table~\ref{tab:main}, our method consistently improves performance across all evaluated backbones.
Averaged over all 144 settings, DBCM reduces MSE by 6.20\% relative to the frozen forecaster, compared with 4.07\% for TAFAS and 2.61\% for PETSA, corresponding to about 1.5$\times$ the gain of the strongest prior method.
This gap widens at longer horizons (4.24\% vs.\ 2.83\% at $H=96$ and 8.87\% vs.\ 5.46\% at $H=720$ for DBCM and TAFAS, respectively).
The improvement is particularly pronounced on the Exchange Rate dataset, which exhibits the strongest non-stationarity according to the ADF statistics in Appendix~\ref{app:data_description}.
For example, under DLinear with $H=720$, the error decreases from 0.8383 to 0.7066 (15.7\% improvement), whereas TAFAS and PETSA achieve only 0.9\% and 3.3\%, respectively.
Similarly, under PatchTST at $H=720$, our method reduces the error from 0.8814 to 0.6609 (25.0\%), while TAFAS and PETSA show negligible gains of 0.02\% and 0.00\%, respectively.
DBCM surpasses channel-wise adaptation methods, indicating that modeling shared and residual components with lightweight nonlinear gating is sufficient for effective adaptation.

\subsection{Scalability Across Model and Data Scales}

\begin{wraptable}{r}{0.5\textwidth}
\vspace{-13pt}
\centering
\caption{\textbf{Scalability.} Average test MSE on Electricity (over four horizons) and on TSFMs (over the four ETT datasets at \(H=96\)).}
\label{tab:scalability}
\setlength{\tabcolsep}{3pt}
\renewcommand{\arraystretch}{0.8}
\small
\begin{tabular}{lcccc}
\toprule
 & Baseline & TAFAS & PETSA & DBCM \\
\midrule
\multicolumn{5}{l}{\textit{Electricity}} \\
iTransformer & .1828 & \underline{.1770} & .1798 & \textbf{.1744} \\
DLinear      & .2149 & \underline{.2093} & .2118 & \textbf{.2063} \\
FreTS        & .1978 & \underline{.1936} & .1956 & \textbf{.1924} \\
\midrule
\multicolumn{5}{l}{\textit{TSFMs}} \\
Moirai    & .5039 & \underline{.3470} & .3507 & \textbf{.3238} \\
Chronos   & .6320 & \underline{.3349} & .3504 & \textbf{.3176} \\
Chronos-2 & .3943 & \underline{.3214} & .3235 & \textbf{.3089} \\
\bottomrule
\end{tabular}
\vspace{-10pt}
\end{wraptable}

We further evaluate DBCM on the high-dimensional Electricity dataset and on three TSFMs, Moirai, Chronos, and Chronos-2.
Table~\ref{tab:scalability} summarizes the results, and the full results are provided in Appendix~\ref{app:scalability}.
On Electricity, DBCM achieves the best performance across all backbones and prediction horizons, with larger gains at longer horizons: for iTransformer at \(H=720\), it improves the baseline by 5.45\%, compared with 2.33\% for TAFAS and 0.83\% for PETSA.
On TSFMs, DBCM also achieves the best performance on all datasets, indicating that the proposed calibration is not tied to a specific foundation model family and can complement large pre-trained forecasters under test-time distribution shifts.

\subsection{Ablation Study}
\label{sec:ablation}

\begin{figure}[t]
    \centering
    \begin{subfigure}[t]{0.58\linewidth}
        \centering
        \includegraphics[width=\linewidth]{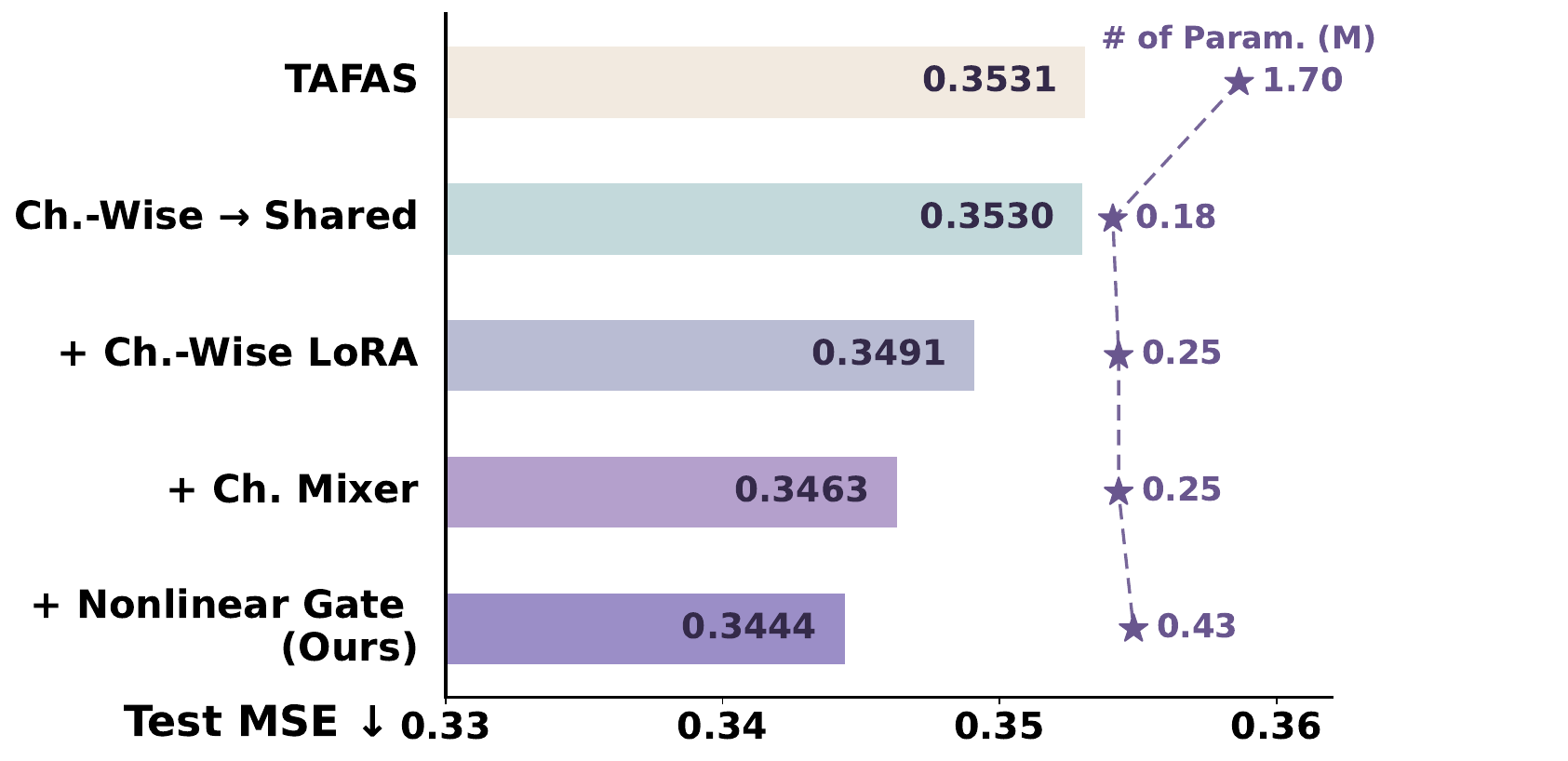}
        \caption{Component ablation.}
        \label{fig:component_ablation}
    \end{subfigure}
    \hfill
    \begin{subfigure}[t]{0.40\linewidth}
        \centering
        \includegraphics[width=\linewidth]{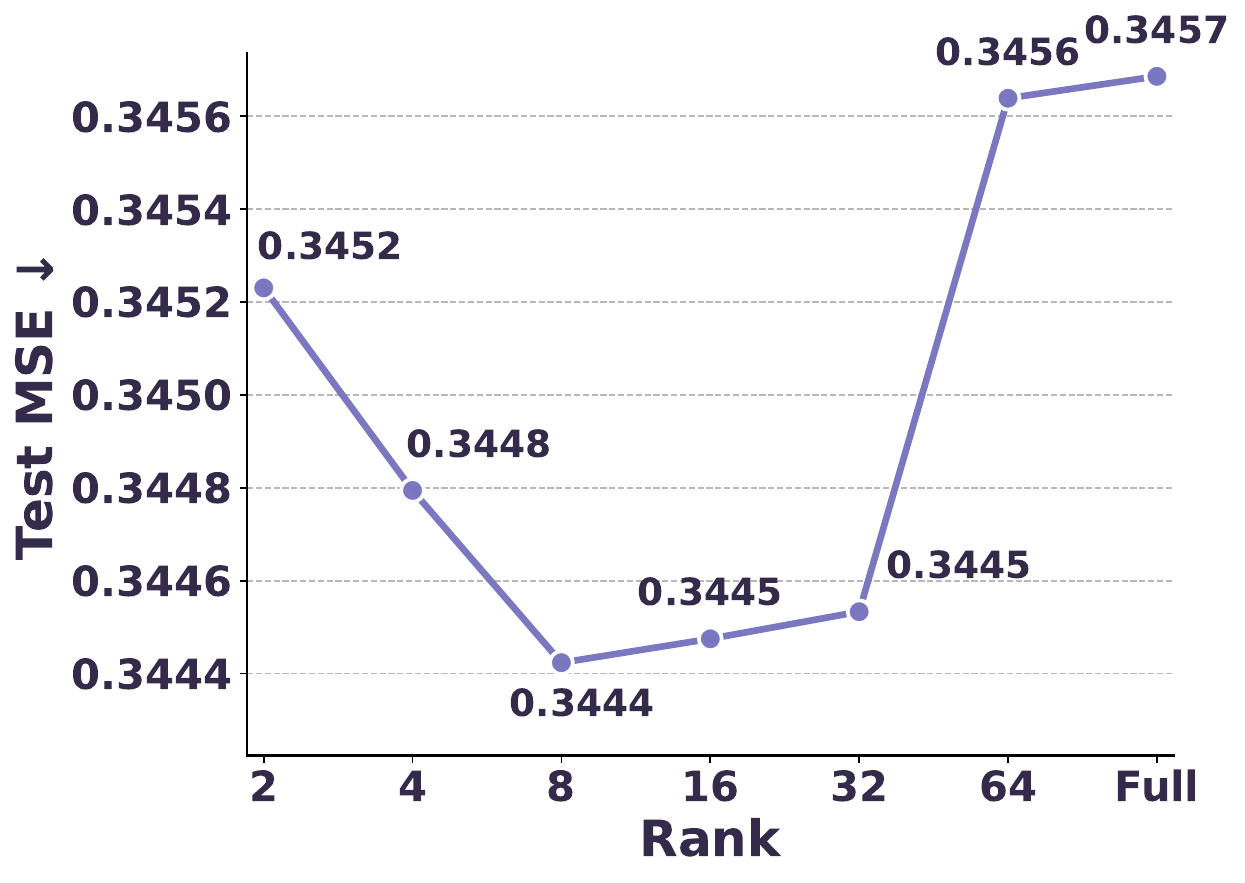}
        \caption{Effect of LoRA rank.}
        \label{fig:rank_ablation}
    \end{subfigure}
    \caption{\textbf{Ablation Study.}
    (a) Component ablation with test MSE (across datasets and horizons) and parameter counts under incremental component addition.
    (b) Effect of LoRA rank on test MSE.}
\end{figure}

\paragraph{Ablation on Components.}
We conduct an ablation study to evaluate each component.
Using iTransformer as the source forecaster, we report test MSE averaged over all six datasets, four forecasting horizons, and three random seeds.
Starting from TAFAS, we first replace its channel-wise linear adapter with a single shared adapter, and then successively add the channel-wise LoRA adapter, the linear channel mixer, and the nonlinear gate, where the final variant corresponds to DBCM.
As shown in Figure~\ref{fig:component_ablation}, replacing the channel-wise adapter in TAFAS with a shared adapter achieves comparable performance with substantially fewer parameters, supporting the redundancy of fully independent channel-wise parameterization suggested by the channel-space analysis in Section~\ref{sec:analysis}, where residuals are concentrated in shared channel directions and TAFAS-Shared aligns with the residual subspace better than TAFAS.
Adding a channel-wise low-rank adapter further improves performance with only a modest parameter increase, complementing the restrictiveness of purely shared adaptation by capturing the residual variation outside the dominant shared directions observed in Section~\ref{sec:analysis}.
The linear channel mixer brings additional gains by providing cross-channel context to the subsequent correction branches, and the nonlinear gating mechanism further refines the correction through input-dependent modulation.
Overall, our final design outperforms TAFAS while using fewer parameters.
Since increasing the LoRA rank beyond 8 does not improve performance (Figure~\ref{fig:rank_ablation}) and the final design uses fewer parameters than TAFAS, these gains are not explained by added capacity alone.
We further show in Appendix~\ref{app:input_output} that input- and output-side DBCM each improve over the source forecaster, and applying both yields the best performance.

\paragraph{Ablation on LoRA Rank.}
We further analyze the effect of the LoRA rank in Figure~\ref{fig:rank_ablation}.
Performance improves as the rank increases from 2 to 8, indicating that a minimal level of channel-specific flexibility is required to complement the shared correction branch.
However, increasing the rank beyond 8 yields no further improvement, and excessively large settings, such as 64 or full-rank adaptation, even degrade performance.

\section{Conclusion}
In this work, we examine the channel-space structure of forecast residuals, which serve as the direct supervision signal for TSF-TTA.
Our analysis shows that these residuals are concentrated along a small number of shared channel directions, and that adapter corrections are well aligned with this residual structure.
Since this structure suggests that fully independent channel-wise parameterization is redundant, we introduce DBCM, which shares adaptation capacity across channels while retaining limited channel-specific flexibility through a lightweight low-rank branch.
More broadly, our results suggest that sharing adaptation capacity across channels while retaining low-rank channel-specific flexibility provides an effective inductive bias for TSF-TTA, rather than assuming fully independent channel-wise corrections.
Since the amount of residual variation can differ across datasets, an important direction for future work is to adaptively determine the channel-wise branch capacity based on these differences.

\paragraph{Limitations.}
While our analysis focuses on the output space, including forecast residuals and output-side corrections, it does not explicitly characterize input-side dynamics, a limitation shared with prior TSF-TTA methods.
The input-side DBCM is nevertheless trained with the same objective on the final calibrated prediction, and using it alone already improves forecasting performance (Appendix~\ref{app:input_output}).
A direct analysis of input-level adaptation remains future work.

\subsection*{Ethics statement}
This work proposes a test-time adaptation method for time series forecasting and uses only publicly available benchmark datasets that contain no personally identifiable information.
We are not aware of any ethical concerns associated with this work.

\bibliography{main}
\bibliographystyle{iclr2027_conference}

\newpage

\appendix

\counterwithin{figure}{section}
\counterwithin{table}{section}
\counterwithin{equation}{section}
\renewcommand{\thetable}{\thesection.\arabic{table}}
\renewcommand{\thefigure}{\thesection.\arabic{figure}}
\renewcommand{\theequation}{\thesection.\arabic{equation}}

\section{Additional Analysis}
\label{app:additional_analysis}

\subsection{Effect of Channel Mixing on Residual Structure.}
\label{app:channel_mixer}
\begin{figure}[!ht]
    \centering
    \includegraphics[width=1.0\columnwidth]{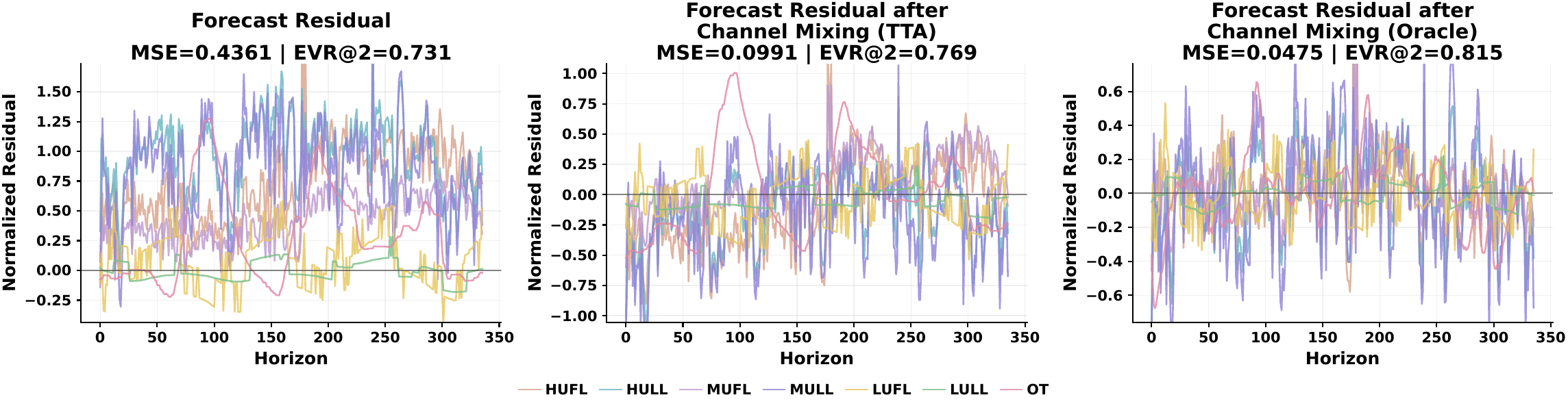}
    \vspace{-5pt}
    \caption{\textbf{Effect of Channel Mixing on Residual Structure.}
    Visualization of forecast residuals for the same test window, from left to right: the original forecast residual, the residual after applying the learned channel mixer adapted during TSF-TTA, and the residual after applying an oracle channel mixer optimized for that window.
    MSE and \(\mathrm{EVR}@2\) are reported in each panel.}
    \vspace{-8pt}
    \label{fig:channel_mixer_residual}
\end{figure}

We further analyze whether the channel mixer indeed makes the subsequent correction problem more structured.
Using iTransformer as the source forecaster, Figure~\ref{fig:channel_mixer_residual} compares the forecast residuals obtained under three settings on the same test window of ETTm2: the original prediction without channel mixing, the prediction obtained with the learned channel mixer adapted under the TSF-TTA setting, and the prediction obtained with an oracle channel mixer optimized for that window to minimize the magnitude of the forecast residual.
The oracle result serves as an upper-bound reference, showing that applying channel mixing before correction can substantially reduce the residual magnitude while making the resulting residual more concentrated along a few channel directions.
More importantly, the learned channel mixer follows the same trend in practice: it reduces MSE from \(0.4361\) to \(0.0991\) and increases \(\mathrm{EVR}@2\) from \(0.731\) to \(0.769\), while the oracle mixer further reaches \(0.0475\) MSE and \(0.815\) \(\mathrm{EVR}@2\).
This suggests that the channel mixer acts as a lightweight preconditioning step, reorganizing channel information before correction so that the subsequent adapters face a smaller residual target that is more concentrated along the channel dimension.

\subsection{Effect of the Channel-Wise LoRA Adapter.}
\begin{figure}[!ht]
    \centering
    \includegraphics[width=1.0\columnwidth]{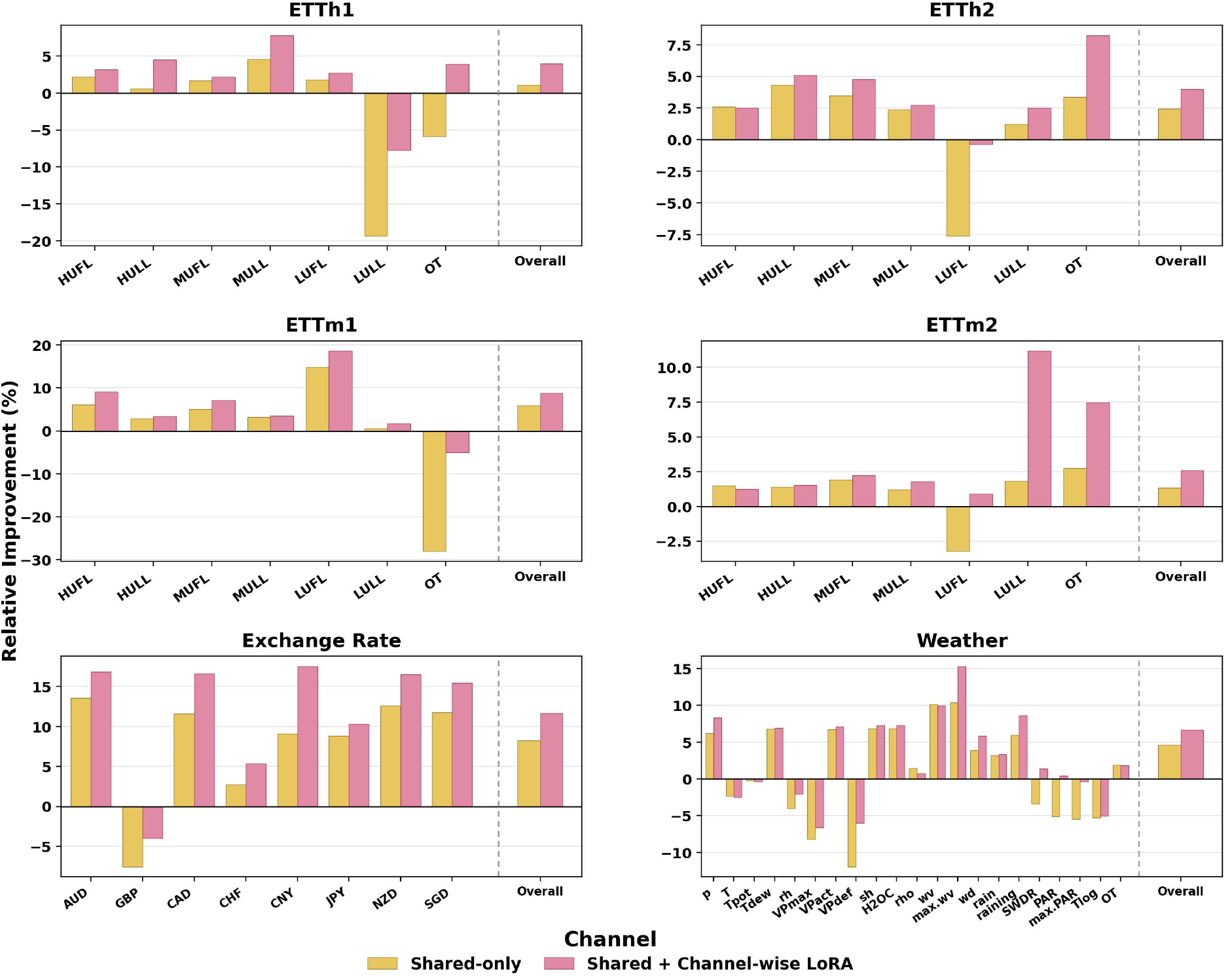}
    \vspace{-5pt}
    \caption{
    \textbf{Channel-Wise Relative Improvement.}
    Channel-wise relative MSE improvement over the iTransformer, comparing the shared-only DBCM variant with the full DBCM using both shared and channel-wise LoRA branches. The overall column reports the relative MSE improvement over all channels.
    }
    \vspace{-8pt}
    \label{fig:relative_improvement}
\end{figure}

We refer to the full model using both the shared branch and the channel-wise LoRA branch as DBCM, and to its variant without the channel-wise LoRA branch as the shared-only variant.
Figure~\ref{fig:relative_improvement} shows the channel-wise relative MSE improvement over the frozen iTransformer for the shared-only variant and DBCM.
The Overall column reports the relative MSE improvement computed over all channels, where DBCM consistently outperforms the shared-only variant across datasets.
Across datasets, DBCM consistently exhibits more favorable channel-wise improvements than the shared-only variant.
In most channels, adding the channel-wise LoRA branch increases the improvement over the source forecaster.
More importantly, for channels where the shared-only variant yields negative relative improvement, DBCM substantially mitigates the degradation and even turns negative improvement into positive improvement in some cases.
This indicates that a purely shared correction can be insufficient or occasionally misaligned for channels whose residual variations deviate from the shared structure, and that the channel-wise branch complements it.
The channel-wise LoRA branch provides lightweight channel-specific flexibility, allowing DBCM to compensate for these remaining variations without relying on fully independent channel-wise adapters.

\subsection{Qualitative Analysis.}
\begin{figure}[!ht]
    \centering
    \includegraphics[width=1.0\columnwidth]{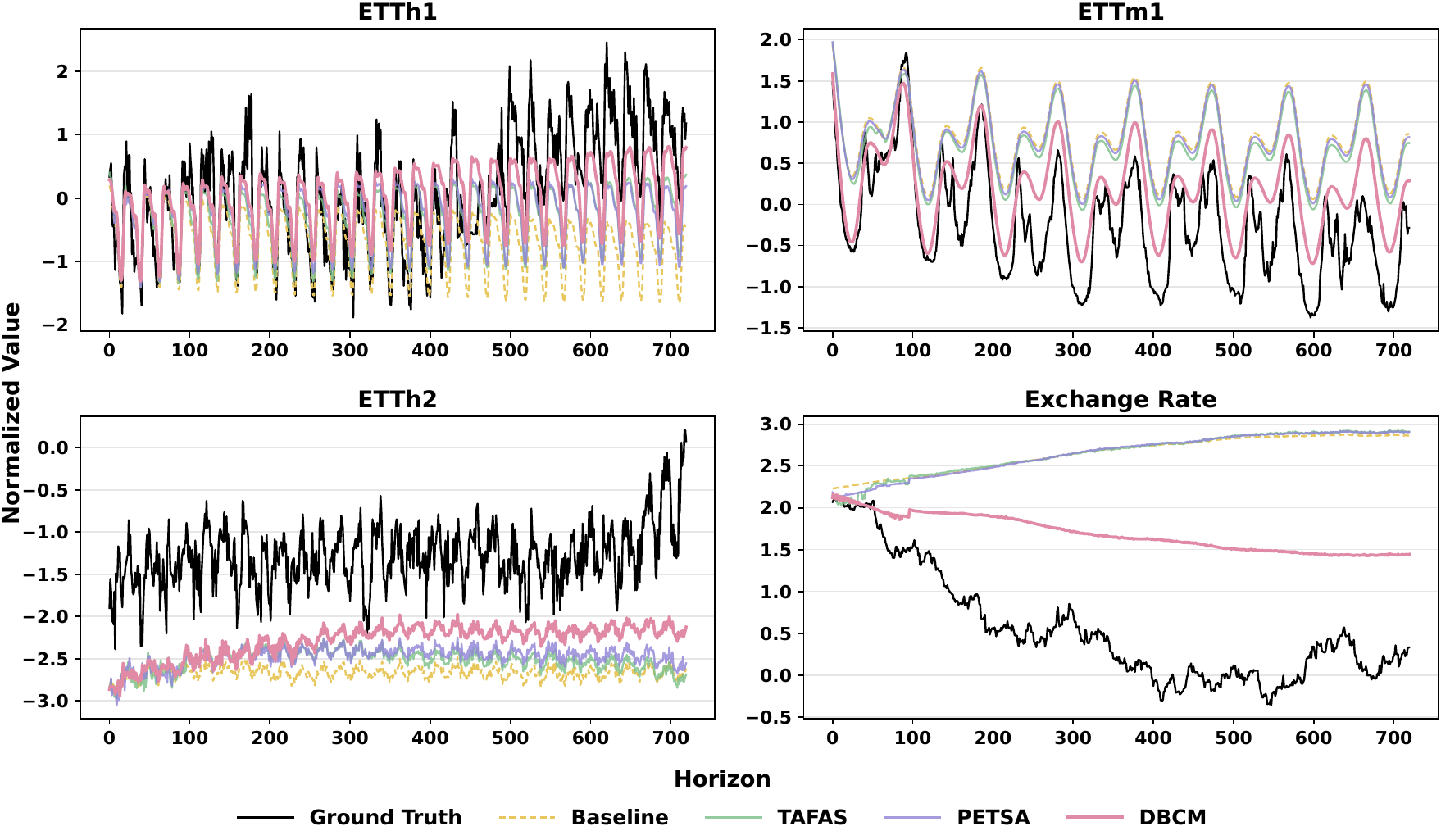}
    \vspace{-5pt}
    \caption{\textbf{Qualitative Analysis.}
    Visualization of forecasting examples on ETTh1, ETTm1, ETTh2, and Exchange Rate using iTransformer as the source forecaster.
    The figure compares the frozen source forecaster and TSF-TTA methods, including TAFAS, PETSA, and DBCM.}
    \vspace{-8pt}
    \label{fig:qualitative_analysis}
\end{figure}

Figure~\ref{fig:qualitative_analysis} presents qualitative forecasting examples under test-time adaptation with iTransformer as the source forecaster.
Across datasets, DBCM produces forecasts that are closer to the ground truth than competing methods, although the observed improvements vary across temporal patterns.
For ETTh1 and ETTm1, DBCM better captures local peaks and follows the overall shape of the ground truth.
In ETTh2, DBCM more accurately matches the scale of the target series, reducing the amplitude mismatch observed in other methods.
For the Exchange Rate dataset, where the baseline and competing methods struggle to capture the trend, DBCM produces predictions that more closely align with the ground truth.
These examples show that DBCM improves qualitative forecasting behavior across different types of test-time shifts, including local peaks, scale changes, and trend shifts.

\subsection{Raw Data Drift Spectra.}
\label{app:raw_drift}
We examine raw data-side drift through channel-wise descriptors, defined as summary statistics that characterize changes across time-series windows.
We consider two reference choices to examine whether the channel-space structure of raw drift depends on how the drift is measured.
The source reference compares each test-window descriptor with the mean descriptor computed over the training windows.
The previous-window reference instead compares each descriptor with that from the immediately preceding test window.

We split the descriptors into two families to distinguish simple statistical shift from changes in temporal structure.
Low-order descriptors capture changes in basic window-level statistics, including mean, variance, and slope.
In contrast, frequency/temporal descriptors capture how the temporal dynamics change, including shifts in spectral energy, dominant periodicity, and autocorrelation.
This separation allows us to examine how simple statistical shifts and richer temporal changes differently shape the channel-space structure of raw drift.
Detailed definitions are provided in Appendix~\ref{app:drift_descriptors}.

As shown in Table~\ref{tab:analysis_drift}, source-referenced low-order drift is relatively concentrated in the channel space, achieving \(\mathrm{EVR}@2=0.854\) and \(\mathrm{Rank}@90=2.89\).
This suggests that simple window-level statistical shifts can share common channel directions.
However, this concentration weakens once frequency and temporal descriptors are included.
For the combined descriptor matrix, \(\mathrm{EVR}@2\) drops to 0.652 under the source reference and 0.596 under the previous-window reference, while \(\mathrm{Rank}@90\) increases to 5.17 and 5.67, respectively.
Thus, the compactness observed in low-order drift does not extend to raw drift as a whole.
In other words, raw drift becomes substantially more heterogeneous once it is characterized beyond simple mean, variance, and trend shifts.
Since TSF-TTA is not optimized to model raw input drift directly, raw data-side drift provides only indirect evidence for adapter design.
We therefore turn to forecast residuals, which serve as the direct supervision signal for TSF-TTA.

\begin{table}[!ht]
\centering
\setlength{\tabcolsep}{3pt}
\renewcommand{\arraystretch}{0.80}
\caption{
\textbf{Raw Data Drift Spectra.}
\(\mathrm{EVR}@k\): top-\(k\) explained variance ratio; \(\mathrm{Rank}@90\): minimum rank explaining 90\% variance.
}
\label{tab:analysis_drift}
\begin{tabular}{lcc}
\toprule
\textbf{Signal} & \(\mathrm{EVR}@2\) & \(\mathrm{Rank}@90\) \\
\midrule
Source, low-order desc.      & 0.854 & 2.89 \\
Source, freq./temporal desc. & 0.728 & 4.11 \\
Source, all desc.            & 0.652 & 5.17 \\
Previous, all desc.          & 0.596 & 5.67 \\
\bottomrule
\end{tabular}
\vspace{-8pt}
\end{table}

\subsection{Dataset-Wise Residual--Correction Alignment.}
\label{app:dataset_wise_analysis}
Table~\ref{tab:dataset_residual_correction_alignment} compares how well the corrections generated by adapters align with the channel-space subspaces of forecast residuals.
We measure the projection similarity between the top-2 subspaces of residuals and corrections for each dataset.
DBCM achieves the highest average alignment score of 0.774, outperforming TAFAS, TAFAS-Shared, and PETSA, and attains the highest alignment on all six datasets.
The improvement is especially pronounced on Weather, which has the largest number of channels, where the corrections of DBCM remain closely aligned with the leading residual subspace.
Moreover, TAFAS-Shared attains higher alignment than TAFAS on average and on five of the six datasets, further supporting that residual corrections are not fully channel-independent and can be modeled through shared channel-space structure.

\begin{table}[!ht]
\centering
\small
\setlength{\tabcolsep}{5pt}
\renewcommand{\arraystretch}{0.85}
\caption{
\textbf{Dataset-Wise Residual--Correction Alignment.}
Projection similarity between the top-2 channel subspaces of forecast residuals and adapter corrections.
}
\label{tab:dataset_residual_correction_alignment}
\begin{tabular}{lcccc}
\toprule
\textbf{Dataset} & \textbf{TAFAS} & \textbf{TAFAS-Shared} & \textbf{PETSA} & \textbf{DBCM} \\
\midrule
ETTh1         & 0.772 & 0.812 & 0.649 & 0.874 \\
ETTh2         & 0.648 & 0.676 & 0.679 & 0.684 \\
ETTm1         & 0.876 & 0.893 & 0.824 & 0.897 \\
ETTm2         & 0.707 & 0.858 & 0.787 & 0.875 \\
Exchange      & 0.457 & 0.554 & 0.533 & 0.603 \\
Weather       & 0.566 & 0.543 & 0.521 & 0.712 \\
\midrule
Average       & 0.671 & 0.723 & 0.666 & 0.774 \\
\bottomrule
\end{tabular}
\end{table}

\subsection{Residual Spectrum on High-Dimensional Data.}
\label{app:elec_spectrum}
Since parameter sharing becomes more relevant as the number of channels grows, we further examine whether the channel-space concentration of forecast residuals persists on the high-dimensional Electricity dataset with 321 channels.
Table~\ref{tab:elec_spectrum} compares Electricity with Weather, the dataset with the most channels in Table~\ref{tab:analysis_spectrum}, using the same centered channel-space metrics.
On Electricity, the top 30\% of channel directions explain 83.2\% of the residual variance, whereas isotropic residuals would yield only 30.2\%, and 90\% of the variance is explained by 45.3\% of the channel directions, compared with 90.0\% for isotropic residuals.
Although the concentration is weaker than on Weather, the residuals remain far from isotropic, indicating that the shared channel-space structure persists in high-dimensional settings.

\begin{table}[!ht]
\centering
\small
\setlength{\tabcolsep}{6pt}
\renewcommand{\arraystretch}{0.9}
\caption{\textbf{Residual Spectrum on High-Dimensional Data.} Centered channel-space metrics of forecast residuals, reported in \%.}
\label{tab:elec_spectrum}
\begin{tabular}{lccc}
\toprule
\textbf{Dataset} & \(C\) & \(\mathrm{EVR}@30\%\) & \(\mathrm{Rank}@90/C\) \\
\midrule
Weather     & 21  & 98.2 & 27.4 \\
Electricity & 321 & 83.2 & 45.3 \\
\bottomrule
\end{tabular}
\end{table}

\FloatBarrier
\section{Algorithm of DBCM}
\label{app:algorithm}
\begin{lstlisting}[style=pytorchstyle, caption={PyTorch-style pseudocode for DBCM}, label={alg:ours}]
# L, H: lookback window length and forecasting horizon
# forecaster: frozen source forecaster
# C_in, C_out: input- and output-side DBCM

forecaster.requires_grad_(False)
C_in, C_out = Calibration(L), Calibration(H)

test_x, test_y, period = [], [], None

# test_gt: H future steps, used only once they are revealed
for test_input, test_gt in test_loader:
    # start a new test mini-batch
    if period is None:
        period = PAAS(test_input)

    test_x.append(test_input)
    test_y.append(test_gt)
    # wait until period + 1 windows have arrived
    if len(test_x) < period + 1:
        continue

    test_x = torch.stack(test_x)
    test_y = torch.stack(test_y)

    # (1) update with partially observed ground truth
    pred_cali = C_out(forecaster(C_in(test_x)))
    loss_p = compute_partial_loss(pred_cali, test_y, period)
    optimizer.zero_grad()
    loss_p.backward()
    optimizer.step()

    # (2) update with fully observed ground truth
    if full_gt_available:
        pred_full = C_out(forecaster(C_in(full_input)))
        loss_f = MSE(pred_full, full_gt)
        optimizer.zero_grad()
        loss_f.backward()
        optimizer.step()

    # (3) adjust predictions for steps not yet observed
    with torch.no_grad():
        pred_adj = C_out(forecaster(C_in(test_x)))
        for i in range(period):
            pred_cali[i, period-i:] = pred_adj[i, period-i:]

    test_x, test_y, period = [], [], None

class Calibration(nn.Module):
    # Z: (T, C), where T = L (input) or H (output)
    def forward(self, Z):
        # linear channel mixer
        Z_mix = Z + Z @ W_mix

        # shared adapter + channel-wise LoRA adapter
        A_sh = W_sh @ Z_mix
        A_cw = stack([(U[c] @ V[c]) @ Z_mix[:, c]
                      for c in range(C)], dim=-1)
        C_base = A_sh + A_cw + b

        # nonlinear gated modulation
        G = sigmoid(W_gate @ Z_mix + b_gate[:, None])
        C_gate = C_base + C_base * G

        # residual path with channel-wise outer gate
        return Z + tanh(g) * C_gate

# test_input: (L, C)
def PAAS(test_input):
    x = test_input - test_input.mean(dim=0)
    amplitude = abs(fft(x, dim=0))
    # dominant frequency of the highest-energy channel
    ch = (amplitude ** 2).sum(dim=0).argmax()
    freq = amplitude[:, ch].argmax()
    return x.shape[0] // freq

def compute_partial_loss(pred_cali, test_y, period):
    B, H = pred_cali.shape[:2]
    # index of the latest window in the batch
    p = min(period, B - 1)
    loss, valid_count = 0.0, 0
    for i in range(p):
        # steps of window i already revealed
        obs_len = min(p - i, H)
        pred_i = pred_cali[i, :obs_len]
        gt_i = test_y[i, :obs_len]
        loss += squared_error(pred_i, gt_i).sum()
        valid_count += pred_i.numel()
    return loss / valid_count
\end{lstlisting}

\section{Additional Experiments}
\label{app:additional_exp}
\subsection{Full Results on Electricity and TSFMs}
\label{app:scalability}

Table~\ref{tab:electricity} reports the results on the Electricity dataset, which contains 321 channels, substantially more than the six main benchmarks.
DBCM achieves the best performance across all backbones and prediction horizons, and the improvement becomes more pronounced at longer horizons.
For iTransformer at \(H=720\), TAFAS and PETSA improve the baseline by only 2.33\% and 0.83\%, respectively, whereas DBCM achieves a larger gain of 5.45\%.
A similar trend is observed at \(H=336\), where TAFAS and PETSA yield improvements of 2.61\% and 0.74\%, whereas DBCM achieves 3.34\%.
These results indicate that existing methods become less effective as the forecasting horizon increases, while DBCM continues to deliver stable improvements in high-dimensional multivariate settings.

Table~\ref{tab:chronos} reports the per-dataset results on Moirai~\citep{woo2024moirai}, Chronos~\citep{ansari2024chronos}, and Chronos-2~\citep{ansari2025chronos2} at \(H=96\).
DBCM achieves the best performance on all four ETT datasets for all three TSFMs, outperforming both TAFAS and PETSA.
The gains are particularly large on ETTm1, where the frozen TSFMs exhibit large errors: DBCM reduces the MSE of Moirai from 1.0543 to 0.4010 and that of Chronos from 1.5167 to 0.3862, compared with 0.4720 and 0.4252 for TAFAS and 0.4708 and 0.4505 for PETSA.
These results indicate that the proposed adaptation mechanism is not tied to a specific foundation model family, but can complement diverse pre-trained TSFMs by calibrating their predictions under test-time distribution shifts.

\begin{table}[!ht]
\centering
\caption{\textbf{Prediction Performance Comparison on the Electricity Dataset.} Test MSE is reported for iTransformer, DLinear and FreTS across four prediction horizons.}
\label{tab:electricity}
\setlength{\tabcolsep}{3pt}
\renewcommand{\arraystretch}{0.80}
\setlength{\aboverulesep}{1.5pt}
\setlength{\belowrulesep}{1.5pt}
\setlength{\abovetopsep}{1.5pt}
\setlength{\belowbottomsep}{1.5pt}
\setlength{\cmidrulesep}{1.5pt}
\small
{
\resizebox{\textwidth}{!}{
\begin{tabular}{ c c c c c c c c c c c c c}
\toprule
 & \multicolumn{4}{c}{\textbf{iTransformer}} & \multicolumn{4}{c}{\textbf{DLinear}} & \multicolumn{4}{c}{\textbf{FreTS}} \\
  \cmidrule(lr){2-5}\cmidrule(lr){6-9}\cmidrule(lr){10-13}
 & Baseline & TAFAS & PETSA & DBCM & Baseline & TAFAS & PETSA & DBCM & Baseline & TAFAS & PETSA & DBCM \\
\midrule
 96 & .1452 & \underline{.1426} & .1446 & \textbf{.1415} & .1974 & \underline{.1897} & .1924 & \textbf{.1844} & .1745 & \underline{.1682} & .1704 & \textbf{.1663} \\
 192 & .1692 & \underline{.1588} & .1612 & \textbf{.1584} & .1971 & \underline{.1907} & .1930 & \textbf{.1886} & .1814 & \underline{.1767} & .1782 & \textbf{.1766} \\
 336 & .1765 & \underline{.1719} & .1752 & \textbf{.1706} & .2117 & \underline{.2059} & .2089 & \textbf{.2051} & .1973 & \underline{.1931} & .1957 & \textbf{.1923}  \\
 720 & .2402 & \underline{.2346} & .2382 & \textbf{.2271} & .2534 & \underline{.2508} & .2530 & \textbf{.2471} & .2380 & \underline{.2363} & .2379 & \textbf{.2345} \\
 \midrule
 Avg. & .1828 & \underline{.1770} & .1798 & \textbf{.1744} & .2149 & \underline{.2093} & .2118 & \textbf{.2063} & .1978 & \underline{.1936} & .1956 & \textbf{.1924} \\
\bottomrule
\end{tabular}
}
}
\end{table}

\begin{table}[!ht]
\centering
\caption{\textbf{TSF-TTA on TSFMs.} Test MSE of TSF-TTA methods evaluated on Moirai, Chronos, and Chronos-2 for horizon \(H=96\).}
\setlength{\tabcolsep}{3pt}
\renewcommand{\arraystretch}{0.75}
\label{tab:chronos}
\begin{tabular}{lcccc}
\toprule
 & ETTh1 & ETTh2 & ETTm1 & ETTm2 \\
\midrule
Moirai & .4917 & .2650 & 1.0543 & .2044 \\
+ TAFAS & \underline{.4740} & \underline{.2557} & .4720 & \underline{.1862} \\
+ PETSA & .4816 & .2628 & \underline{.4708} & .1875 \\
+ DBCM & \textbf{.4629} & \textbf{.2524} & \textbf{.4010} & \textbf{.1788}  \\
\midrule
Chronos & .5316 & .2553 & 1.5167 & .2242 \\
+ TAFAS & \underline{.4811} & \underline{.2467} & \underline{.4252} & .1866 \\
+ PETSA & .5147 & .2504 & .4505 & \underline{.1861} \\
+ DBCM & \textbf{.4602} & \textbf{.2461} & \textbf{.3862} & \textbf{.1779}  \\
\midrule
Chronos-2 & .4904 & .2485 & .6544 & .1840 \\
+ TAFAS & \underline{.4643} & \underline{.2408} & \underline{.4109} & .1698 \\
+ PETSA & .4687 & .2419 & .4148 & \underline{.1686} \\
+ DBCM & \textbf{.4540} & \textbf{.2385} & \textbf{.3772} & \textbf{.1657}  \\
\bottomrule
\end{tabular}
\end{table}

\subsection{Comparison with DynaTTA}

We report the complete prediction performance comparison with DynaTTA~\citep{grover2025shiftaware}.
DynaTTA involves 15 hyperparameters; for comparison, we follow the configuration used in COSA~\citep{im2026cosa}.
We also note that the official DynaTTA implementation has reproducibility issues.
As shown in Table~\ref{tab:dynatta}, DynaTTA exhibits unstable adaptation behavior across datasets and backbone models, often increasing the forecasting error compared to the frozen baseline.
In contrast, DBCM consistently achieves the best performance across all evaluated settings.
These results suggest that dynamically adjusting adaptation behavior through many hyperparameters can be difficult to stabilize, while DBCM provides more reliable test-time adaptation across diverse forecasting architectures and distribution shifts.

\begin{table}[!ht]
\centering
\caption{\textbf{Prediction Performance Comparison with DynaTTA.} Test MSE is reported for three backbone models on six datasets with four prediction horizons. Bold entries indicate the best performance among methods, while underlined entries denote the second-best.}
\label{tab:dynatta}
\setlength{\tabcolsep}{3pt}
\renewcommand{\arraystretch}{0.70}
\setlength{\aboverulesep}{1.5pt}
\setlength{\belowrulesep}{1.5pt}
\setlength{\abovetopsep}{1.5pt}
\setlength{\belowbottomsep}{1.5pt}
\setlength{\cmidrulesep}{1.5pt}
\scriptsize
{
\resizebox{\textwidth}{!}{%
\begin{tabular}{c c c c c c c c c c c}
\toprule
\multicolumn{2}{c}{}
& \multicolumn{3}{c}{\textbf{iTransformer}}
& \multicolumn{3}{c}{\textbf{DLinear}}
& \multicolumn{3}{c}{\textbf{FreTS}} \\
\cmidrule(lr){3-5}\cmidrule(lr){6-8}\cmidrule(lr){9-11}
\multicolumn{2}{c}{}
& Baseline & DynaTTA & DBCM
& Baseline & DynaTTA & DBCM
& Baseline & DynaTTA & DBCM \\
\midrule
\multirow{4}{*}{\rotatebox[origin=c]{90}{\textbf{ETTh1}}}
& 96  & \underline{.4349} & .5253 & \textbf{.4228} & \underline{.4510} & .4632 & \textbf{.4318} & \underline{.4360} & .5199 & \textbf{.4242} \\
& 192 & \underline{.4937} & .5608 & \textbf{.4805} & \underline{.5042} & .5161 & \textbf{.4835} & \underline{.4886} & .5505 & \textbf{.4764} \\
& 336 & \underline{.5406} & .5979 & \textbf{.5322} & \underline{.5508} & .5687 & \textbf{.5373} & \underline{.5466} & .6277 & \textbf{.5354} \\
& 720 & \underline{.6914} & .7576 & \textbf{.6298} & \underline{.7019} & .7080 & \textbf{.6487} & \underline{.7145} & .9315 & \textbf{.6565} \\
\midrule
\multirow{4}{*}{\rotatebox[origin=c]{90}{\textbf{ETTh2}}}
& 96  & \underline{.2430} & .3498 & \textbf{.2388} & \underline{.2294} & .2305 & \textbf{.2268} & \underline{.2376} & .2523 & \textbf{.2347} \\
& 192 & \underline{.2973} & .3669 & \textbf{.2914} & \underline{.2835} & .2871 & \textbf{.2794} & \underline{.2838} & .3111 & \textbf{.2761} \\
& 336 & \underline{.3448} & .4136 & \textbf{.3301} & \underline{.3250} & .3370 & \textbf{.3152} & \underline{.3184} & .3607 & \textbf{.3048} \\
& 720 & \underline{.4258} & .4900 & \textbf{.3939} & \underline{.4149} & .4361 & \textbf{.3820} & \underline{.3951} & .4225 & \textbf{.3653} \\
\midrule
\multirow{4}{*}{\rotatebox[origin=c]{90}{\textbf{ETTm1}}}
& 96  & \underline{.3763} & .8006 & \textbf{.3415} & \underline{.3705} & .3910 & \textbf{.3420} & \underline{.3618} & .4564 & \textbf{.3420} \\
& 192 & \underline{.4371} & .8724 & \textbf{.4038} & \underline{.4420} & .4995 & \textbf{.4056} & \underline{.4304} & .5729 & \textbf{.4060} \\
& 336 & \underline{.5029} & .9139 & \textbf{.4628} & \underline{.5172} & .5983 & \textbf{.4626} & \underline{.4928} & .6636 & \textbf{.4618} \\
& 720 & \underline{.5713} & .7554 & \textbf{.5276} & \underline{.5927} & .6570 & \textbf{.5321} & \underline{.5609} & .7543 & \textbf{.5274} \\
\midrule
\multirow{4}{*}{\rotatebox[origin=c]{90}{\textbf{ETTm2}}}
& 96  & \underline{.1555} & .1903 & \textbf{.1536} & \underline{.1589} & .1632 & \textbf{.1573} & \underline{.1563} & .2023 & \textbf{.1536} \\
& 192 & \underline{.1938} & .2237 & \textbf{.1923} & \underline{.1929} & .2067 & \textbf{.1907} & \underline{.1914} & .2117 & \textbf{.1880} \\
& 336 & \underline{.2345} & .3038 & \textbf{.2312} & \underline{.2325} & .2970 & \textbf{.2289} & \underline{.2310} & .2986 & \textbf{.2259} \\
& 720 & \underline{.3101} & .4069 & \textbf{.2990} & \underline{.3061} & .3827 & \textbf{.2956} & \underline{.3026} & .4319 & \textbf{.2930} \\
\midrule
\multirow{4}{*}{\rotatebox[origin=c]{90}{\parbox{0.95cm}{\centering\textbf{Exchange}}}}
& 96  & \underline{.0916} & .1097 & \textbf{.0858} & \underline{.0811} & .0842 & \textbf{.0789} & \underline{.0827} & .1472 & \textbf{.0794} \\
& 192 & \underline{.1914} & .2310 & \textbf{.1770} & \underline{.1708} & .1846 & \textbf{.1626} & \underline{.1787} & .2558 & \textbf{.1659} \\
& 336 & .3375 & \underline{.3100} & \textbf{.2844} & .3205 & \underline{.2965} & \textbf{.2642} & \underline{.3265} & .3885 & \textbf{.2659} \\
& 720 & \underline{.9433} & .9968 & \textbf{.8182} & .8383 & \underline{.8381} & \textbf{.7066} & .8251 & \underline{.8242} & \textbf{.6700} \\
\midrule
\multirow{4}{*}{\rotatebox[origin=c]{90}{\textbf{Weather}}}
& 96  & \underline{.1719} & .2287 & \textbf{.1630} & .1936 & \underline{.1839} & \textbf{.1732} & \underline{.1809} & .1903 & \textbf{.1677} \\
& 192 & \underline{.2233} & .3659 & \textbf{.2096} & \underline{.2383} & .2963 & \textbf{.2194} & \underline{.2266} & .3167 & \textbf{.2101} \\
& 336 & \underline{.2800} & .6478 & \textbf{.2608} & \underline{.2913} & .5204 & \textbf{.2697} & \underline{.2806} & .3872 & \textbf{.2602} \\
& 720 & \underline{.3576} & .5251 & \textbf{.3360} & \underline{.3643} & .5951 & \textbf{.3391} & \underline{.3557} & .5414 & \textbf{.3338} \\
\midrule
\multicolumn{2}{c}{\textbf{Average}}
& \underline{.3687} & .4977 & \textbf{.3444}
& \underline{.3655} & .4059 & \textbf{.3389}
& \underline{.3585} & .4425 & \textbf{.3343} \\
\midrule
\multicolumn{2}{c}{\textbf{Avg. Rank}}
& \underline{2.04} & 2.96 & \textbf{1.00}
& \underline{2.12} & 2.88 & \textbf{1.00}
& \underline{2.04} & 2.96 & \textbf{1.00} \\
\bottomrule
\end{tabular}
}
}
\end{table}

\subsection{Comparison with COSA}

DBCM follows the TSF-TTA protocol of TAFAS~\citep{kim2025battling} and PETSA~\citep{medeiros2025accurate}.
Under this protocol, the adapter is updated only with observations available at the current test time.
Specifically, adaptation uses partially observed ground truth for the current prediction and, when available, fully observed ground truth from past buffered predictions.
COSA~\citep{im2026cosa}, in contrast, adopts a different adaptation protocol: it first evaluates a test batch and then directly updates the adapter using the ground truth of the same evaluated batch, regardless of whether those target values have been observed at the corresponding test time.
This protocol can expose the adapter to future target values that would not be available in a streaming forecasting scenario, leading to label leakage.

To directly compare DBCM with COSA under the same update strategy, we additionally apply the COSA protocol to DBCM and report the results in Table~\ref{tab:cosa}.
Although this protocol is not used in our main experiments due to the aforementioned leakage issue, this comparison isolates the effect of the adaptation module from differences in the update protocol.
COSA provides two variants: COSA-F updates the adapter at a fixed interval, using 48 time steps as the default, while COSA-P uses PAAS from TAFAS~\citep{kim2025battling} to determine the update interval from the periodicity of the input sequence.
Following TAFAS, DBCM also uses PAAS by default to construct test mini-batches for adaptation.
Under COSA's protocol, DBCM achieves the best average performance across all three evaluated backbones.
These results indicate that, even when evaluated under the same update strategy as COSA, DBCM provides more effective correction on average, suggesting that its dual-branch correction structure is beneficial beyond differences in the test-time adaptation protocol.

\begin{table}[!ht]
\centering
\caption{\textbf{Prediction Performance Comparison with COSA.} Test MSE obtained by applying the update protocol used in COSA. Bold entries indicate the best performance among methods, while underlined entries denote the second-best.}
\label{tab:cosa}
\setlength{\tabcolsep}{3pt}
\renewcommand{\arraystretch}{0.70}
\setlength{\aboverulesep}{1.5pt}
\setlength{\belowrulesep}{1.5pt}
\setlength{\abovetopsep}{1.5pt}
\setlength{\belowbottomsep}{1.5pt}
\setlength{\cmidrulesep}{1.5pt}
\scriptsize
{
\resizebox{\textwidth}{!}{%
\begin{tabular}{c c c c c c c c c c c c c c}
\toprule
\multicolumn{2}{c}{} & \multicolumn{4}{c}{\textbf{iTransformer}} & \multicolumn{4}{c}{\textbf{DLinear}} & \multicolumn{4}{c}{\textbf{FreTS}} \\
\cmidrule(lr){3-6}\cmidrule(lr){7-10}\cmidrule(lr){11-14}
\multicolumn{2}{c}{} & Baseline & COSA-F & COSA-P & DBCM & Baseline & COSA-F & COSA-P & DBCM & Baseline & COSA-F & COSA-P & DBCM \\
\midrule
\multirow{4}{*}{\rotatebox[origin=c]{90}{\textbf{ETTh1}}} & 96 & .4349 & \underline{.4325} & .4326 & \textbf{.4272} & .4510 & .4437 & \underline{.4433} & \textbf{.4354} & .4360 & \underline{.4324} & .4326 & \textbf{.4261} \\
 & 192 & .4937 & .4920 & \underline{.4913} & \textbf{.4790} & .5042 & \underline{.4977} & .4997 & \textbf{.4867} & .4886 & \underline{.4875} & .4877 & \textbf{.4769} \\
 & 336 & .5406 & .5388 & \underline{.5322} & \textbf{.5177} & .5508 & .5458 & \underline{.5420} & \textbf{.5274} & .5466 & .5433 & \underline{.5383} & \textbf{.5213} \\
 & 720 & .6914 & .6452 & \underline{.6229} & \textbf{.5738} & .7019 & .6635 & \underline{.6425} & \textbf{.5853} & .7145 & .6768 & \underline{.6573} & \textbf{.6054} \\ \midrule
\multirow{4}{*}{\rotatebox[origin=c]{90}{\textbf{ETTh2}}} & 96 & .2430 & .2418 & \underline{.2408} & \textbf{.2391} & .2294 & .2285 & \underline{.2281} & \textbf{.2273} & .2376 & .2371 & \underline{.2364} & \textbf{.2359} \\
 & 192 & .2973 & \underline{.2955} & \textbf{.2953} & .3019 & \underline{.2835} & \textbf{.2826} & \textbf{.2826} & .2891 & .2838 & \underline{.2829} & \textbf{.2806} & .2851 \\
 & 336 & .3448 & .3429 & \underline{.3422} & \textbf{.3335} & \underline{.3250} & .3266 & .3267 & \textbf{.3178} & .3184 & \underline{.3172} & .3173 & \textbf{.3066} \\
 & 720 & .4258 & \underline{.4053} & .4169 & \textbf{.3818} & .4149 & \underline{.3945} & .4059 & \textbf{.3726} & .3951 & \underline{.3745} & .3839 & \textbf{.3542} \\ \midrule
\multirow{4}{*}{\rotatebox[origin=c]{90}{\textbf{ETTm1}}} & 96 & .3763 & \underline{.3535} & .3589 & \textbf{.3404} & .3705 & \underline{.3486} & .3487 & \textbf{.3428} & .3618 & \underline{.3523} & .3532 & \textbf{.3428} \\
 & 192 & .4371 & \underline{.4184} & .4211 & \textbf{.4058} & .4420 & .4178 & \underline{.4165} & \textbf{.4085} & .4304 & .4199 & \underline{.4193} & \textbf{.4094} \\
 & 336 & .5029 & \underline{.4850} & .4863 & \textbf{.4676} & .5172 & .4877 & \underline{.4853} & \textbf{.4718} & .4928 & .4831 & \underline{.4819} & \textbf{.4655} \\
 & 720 & .5713 & \underline{.5562} & .5601 & \textbf{.5299} & .5927 & \underline{.5581} & .5632 & \textbf{.5370} & .5609 & \underline{.5505} & .5532 & \textbf{.5285} \\ \midrule
\multirow{4}{*}{\rotatebox[origin=c]{90}{\textbf{ETTm2}}} & 96 & .1555 & \underline{.1550} & .1551 & \textbf{.1539} & .1589 & .1589 & \underline{.1588} & \textbf{.1578} & .1563 & \underline{.1556} & \underline{.1556} & \textbf{.1544} \\
 & 192 & \textbf{.1938} & .1948 & \underline{.1941} & .1958 & \textbf{.1929} & .1941 & \textbf{.1929} & \underline{.1936} & \underline{.1914} & .1928 & \textbf{.1913} & \underline{.1914} \\
 & 336 & \underline{.2345} & .2357 & .2353 & \textbf{.2344} & \textbf{.2325} & .2350 & \underline{.2327} & .2365 & .2310 & .2321 & \underline{.2306} & \textbf{.2292} \\
 & 720 & .3101 & \textbf{.3005} & \underline{.3098} & .3296 & .3061 & \textbf{.2965} & \underline{.3042} & .3480 & .3026 & \textbf{.2922} & \underline{.3000} & .3315 \\ \midrule
\multirow{4}{*}{\rotatebox[origin=c]{90}{\parbox{0.95cm}{\centering\textbf{Exchange}}}} & 96 & .0916 & \underline{.0896} & .0905 & \textbf{.0852} & .0811 & \underline{.0800} & .0806 & \textbf{.0786} & .0827 & \underline{.0810} & .0820 & \textbf{.0788} \\
 & 192 & .1914 & \underline{.1813} & .1881 & \textbf{.1730} & .1708 & \underline{.1640} & .1689 & \textbf{.1605} & .1787 & \underline{.1705} & .1760 & \textbf{.1644} \\
 & 336 & .3375 & \underline{.3141} & .3298 & \textbf{.2997} & .3205 & \underline{.2977} & .3129 & \textbf{.2809} & .3265 & \underline{.3031} & .3185 & \textbf{.2801} \\
 & 720 & .9433 & \underline{.7529} & .8907 & \textbf{.5979} & .8383 & \underline{.6512} & .7904 & \textbf{.6170} & .8251 & \underline{.6430} & .7788 & \textbf{.6026} \\ \midrule
\multirow{4}{*}{\rotatebox[origin=c]{90}{\textbf{Weather}}} & 96 & .1719 & \textbf{.1653} & \underline{.1676} & .1682 & .1936 & \underline{.1860} & .1883 & \textbf{.1827} & .1809 & \underline{.1763} & .1768 & \textbf{.1708} \\
 & 192 & .2233 & \underline{.2158} & .2174 & \textbf{.2143} & .2383 & \underline{.2283} & .2296 & \textbf{.2255} & .2266 & \underline{.2196} & .2202 & \textbf{.2126} \\
 & 336 & .2800 & .2720 & \underline{.2718} & \textbf{.2704} & .2913 & .2808 & \underline{.2805} & \textbf{.2762} & .2806 & .2721 & \underline{.2718} & \textbf{.2671} \\
 & 720 & .3576 & \textbf{.3323} & \underline{.3426} & .3437 & .3643 & \underline{.3391} & .3480 & \textbf{.3382} & .3557 & \textbf{.3327} & \underline{.3412} & .3423 \\
\midrule
\multicolumn{2}{c}{\textbf{Average}} & .3687 & \underline{.3507} & .3581 & \textbf{.3360} & .3655 & \underline{.3461} & .3530 & \textbf{.3374} & .3585 & \underline{.3429} & .3494 & \textbf{.3326} \\
\midrule
\multicolumn{2}{c}{\textbf{Avg. Rank}} & 3.71 & \underline{2.25} & 2.50 & \textbf{1.54} & 3.58 & \underline{2.46} & 2.50 & \textbf{1.46} & 3.81 & \underline{2.35} & 2.44 & \textbf{1.40} \\
\bottomrule
\end{tabular}
}
}
\end{table}

\subsection{Comparison with DynaME}
TSF-TTA and online learning for time series forecasting may appear similar, as both exploit sequentially arriving observations after deployment.
However, they differ in their adaptation objectives and deployment constraints.
Online learning broadly aims to continuously update the forecasting system itself from streaming data, which may involve modifying the backbone, prediction head, memory, or ensemble weights.
In contrast, TSF-TTA assumes a pre-trained source forecaster whose learned temporal knowledge should be preserved, and restricts test-time updates to lightweight auxiliary modules that calibrate predictions under distribution shifts using delayed or partially observed feedback. DynaME~\citep{hong2026dynamic} is one of the state-of-the-art methods in the online learning literature for time series forecasting.
We compare our DBCM with DynaME using iTransformer as the source forecaster. In DBCM, we freeze the source forecaster, whereas DynaME updates the forecaster itself.

For a fair comparison, we follow the dataset splits used throughout this paper.
As shown in Table~\ref{tab:dyname}, DBCM achieves the lowest average MSE and the best average rank, reducing the average MSE from 0.3687 to 0.3444 over the baseline and from 0.3614 to 0.3444 over DynaME.
Notably, DBCM consistently reduces the forecasting error over the baseline across all 24 dataset--horizon settings, whereas DynaME occasionally degrades the baseline performance, e.g., on ETTh1, ETTm2, and Exchange Rate.
This suggests that directly updating the forecaster in an online manner can be sensitive to the target stream and may suffer from negative adaptation under certain shifts.
In contrast, DBCM provides more stable improvements by preserving the source forecaster and adapting only the calibration modules.
The advantage of DBCM is particularly clear at the longest horizon, where it achieves the best performance on all six datasets, indicating that structured calibration is especially effective under severe long-range distribution shifts.

\begin{table}[!ht]
\centering
\caption{\textbf{Prediction Performance Comparison with DynaME.} Test MSE is reported for iTransformer on six datasets with four prediction horizons. Bold entries indicate the best performance among methods, while underlined entries denote the second-best.}
\label{tab:dyname}
\renewcommand{\arraystretch}{0.70}
\setlength{\aboverulesep}{2pt}
\setlength{\belowrulesep}{2pt}
\setlength{\abovetopsep}{2pt}
\setlength{\belowbottomsep}{2pt}
\setlength{\cmidrulesep}{2pt}
\begin{tabular}{c c c c c}
\toprule
\multicolumn{2}{c}{}
& \multicolumn{3}{c}{\textbf{iTransformer}} \\
\cmidrule(lr){3-5}
\multicolumn{2}{c}{}
& Baseline & DynaME & DBCM \\
\midrule
\multirow{4}{*}{\rotatebox[origin=c]{90}{\textbf{ETTh1}}}
& 96  & \underline{.4349} & .4370 & \textbf{.4228} \\
& 192 & .4937 & \underline{.4933} & \textbf{.4805} \\
& 336 & \underline{.5406} & .5509 & \textbf{.5322} \\
& 720 & \underline{.6914} & .7008 & \textbf{.6298} \\
\midrule
\multirow{4}{*}{\rotatebox[origin=c]{90}{\textbf{ETTh2}}}
& 96  & .2430 & \textbf{.2379} & \underline{.2388} \\
& 192 & .2973 & \underline{.2934} & \textbf{.2914} \\
& 336 & .3448 & \underline{.3347} & \textbf{.3301} \\
& 720 & \underline{.4258} & .4289 & \textbf{.3939} \\
\midrule
\multirow{4}{*}{\rotatebox[origin=c]{90}{\textbf{ETTm1}}}
& 96  & .3763 & \textbf{.3274} & \underline{.3415} \\
& 192 & .4371 & \textbf{.3941} & \underline{.4038} \\
& 336 & .5029 & \textbf{.4591} & \underline{.4628} \\
& 720 & .5713 & \underline{.5563} & \textbf{.5276} \\
\midrule
\multirow{4}{*}{\rotatebox[origin=c]{90}{\textbf{ETTm2}}}
& 96  & .1555 & \underline{.1538} & \textbf{.1536} \\
& 192 & \underline{.1938} & .1944 & \textbf{.1923} \\
& 336 & \underline{.2345} & .2350 & \textbf{.2312} \\
& 720 & .3101 & \underline{.3093} & \textbf{.2990} \\
\midrule
\multirow{4}{*}{\rotatebox[origin=c]{90}{\parbox{1.32cm}{\centering\textbf{Exchange}}}}
& 96  & \underline{.0916} & .0966 & \textbf{.0858} \\
& 192 & \underline{.1914} & .2004 & \textbf{.1770} \\
& 336 & \underline{.3375} & .3654 & \textbf{.2844} \\
& 720 & .9433 & \underline{.9265} & \textbf{.8182} \\
\midrule
\multirow{4}{*}{\rotatebox[origin=c]{90}{\textbf{Weather}}}
& 96  & .1719 & \textbf{.1622} & \underline{.1630} \\
& 192 & .2233 & \underline{.2113} & \textbf{.2096} \\
& 336 & .2800 & \underline{.2657} & \textbf{.2608} \\
& 720 & .3576 & \underline{.3391} & \textbf{.3360} \\
\midrule
\multicolumn{2}{c}{\textbf{Average}}
& .3687 & \underline{.3614} & \textbf{.3444} \\
\midrule
\multicolumn{2}{c}{\textbf{Avg. Rank}}
& 2.63 & \underline{2.17} & \textbf{1.21} \\
\bottomrule
\end{tabular}
\end{table}

\subsection{Comparison with Normalization Methods}

Normalization methods are widely used to mitigate non-stationarity in time series forecasting. RevIN~\citep{kim2021reversible} removes instance-specific statistics from the input sequence and restores them after forecasting, while DDN~\citep{dai2024ddn} performs dynamic normalization in both the time and frequency domains to better handle non-stationary temporal patterns. Since these methods are designed to alleviate distributional discrepancies in the source forecaster, we further examine whether TSF-TTA provides additional benefits when the source forecaster is equipped with such normalization modules. To this end, we compare TAFAS, PETSA, and DBCM under three settings: without an additional normalization module, with RevIN, and with DDN.

\begin{figure}[!ht]
    \centering
    \includegraphics[width=1.0\columnwidth]{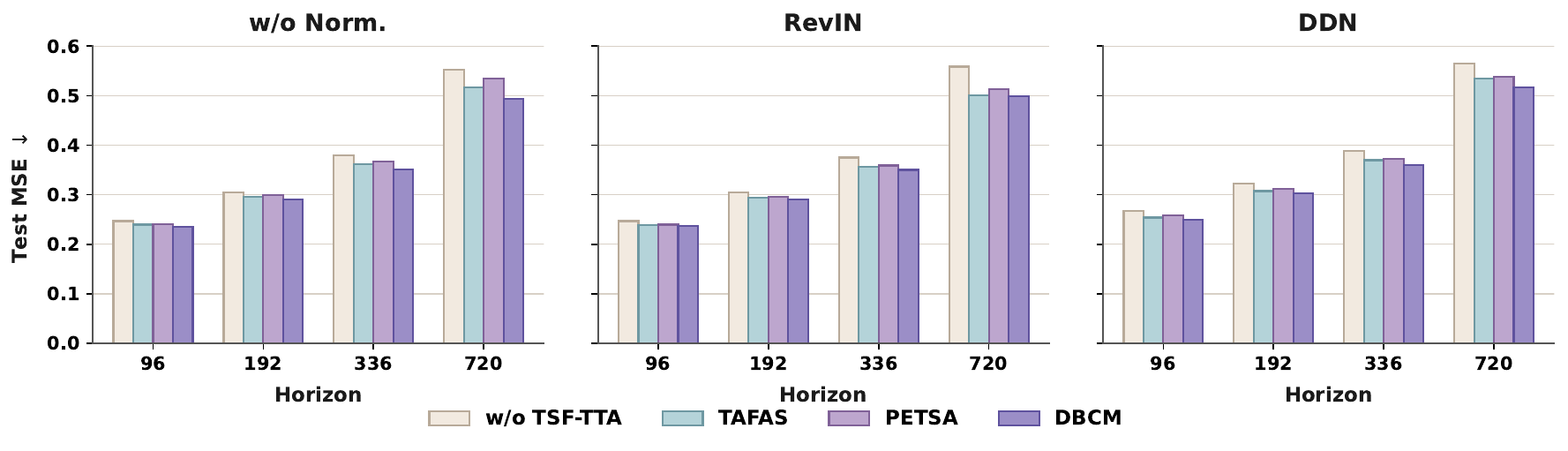}
    \vspace{-5pt}
    \caption{\textbf{Prediction Performance under Different Normalization Settings.}
    Test MSE of the source forecaster without TSF-TTA and with TAFAS, PETSA, or DBCM across prediction horizons.
    The three panels correspond to settings without an additional normalization module (w/o Norm.), with RevIN, and with DDN, respectively.
    For each horizon, results are averaged over six source forecasters and six datasets. Lower values indicate better performance.}
    \vspace{-8pt}
    \label{fig:norm_module}
\end{figure}

Figure~\ref{fig:norm_module} reports the test MSE across prediction horizons, averaged over six source forecasters and six datasets. Across all three normalization settings, applying TSF-TTA consistently reduces the test MSE compared with the corresponding source forecaster without test-time adaptation. This demonstrates that TSF-TTA remains effective even when RevIN or DDN is employed, suggesting that test-time adaptation is complementary to normalization-based non-stationarity handling rather than merely replacing it. Among the compared methods, DBCM achieves the lowest test MSE across all prediction horizons and normalization settings. The absolute MSE reduction generally becomes more pronounced at longer horizons, particularly at \(H=720\), indicating that the proposed dual-branch calibration mechanism effectively corrects residual prediction errors that remain after normalization.

\subsection{Evaluation on TTFBench}

Standard TSF benchmarks often exhibit only mild distribution shifts in their original test splits, which may understate the effectiveness of test-time adaptation under severe deployment-time non-stationarity.
To provide a more challenging stress test, we additionally evaluate on TTFBench, a synthetic-shift benchmark for TSF-TTA.
In this setting, the train and validation splits are kept unchanged, and the source forecaster is trained and selected using these original splits.
Controlled distribution shifts are then injected only into the test split, so that the trained model is evaluated on test streams with harsher non-stationarity than those in the standard benchmark setting.
The injected shifts include piecewise polynomial trends, sinusoidal seasonal components, discrete regime changes, and Gaussian noise with varying intensities.
For reproducibility, we use the TTFBench dataset generated with seed 0.

Table~\ref{tab:ttfbench} reports the forecasting performance on TTFBench.
Overall, DBCM consistently improves robustness under synthesized test-time shifts.
For DLinear and FreTS, DBCM achieves both the lowest average MSE and the best average rank, reducing the average MSE from 0.4068 to 0.3825 and from 0.3999 to 0.3782, respectively.
For iTransformer, DBCM obtains the best average rank of 1.42, although its average MSE is slightly higher than that of TAFAS.
This gap is mainly driven by the Exchange Rate results at the longest prediction horizon, where TAFAS achieves a substantially lower MSE.
In contrast, the average-rank result indicates that DBCM achieves more consistently strong performance across all dataset--horizon pairs.
These results demonstrate that DBCM provides stable and effective test-time adaptation under harsher synthesized distribution shifts.

\begin{table}[!ht]
\centering
\caption{\textbf{Prediction Performance Comparison in TTFBench.} Test MSE evaluated on TTFBench is reported for three source forecasters on six datasets with four prediction horizons. Bold entries indicate the best performance among methods, while underlined entries denote the second-best.
}
\label{tab:ttfbench}
\setlength{\tabcolsep}{3pt}
\renewcommand{\arraystretch}{0.70}
\setlength{\aboverulesep}{1.5pt}
\setlength{\belowrulesep}{1.5pt}
\setlength{\abovetopsep}{1.5pt}
\setlength{\belowbottomsep}{1.5pt}
\setlength{\cmidrulesep}{1.5pt}
\scriptsize
{
\resizebox{\textwidth}{!}{%
\begin{tabular}{c c c c c c c c c c c c c c}
\toprule
\multicolumn{2}{c}{} & \multicolumn{4}{c}{\textbf{iTransformer}} & \multicolumn{4}{c}{\textbf{DLinear}} & \multicolumn{4}{c}{\textbf{FreTS}} \\
\cmidrule(lr){3-6}\cmidrule(lr){7-10}\cmidrule(lr){11-14}
\multicolumn{2}{c}{} & Baseline & TAFAS & PETSA & DBCM & Baseline & TAFAS & PETSA & DBCM & Baseline & TAFAS & PETSA & DBCM \\
\midrule
\multirow{4}{*}{\rotatebox[origin=c]{90}{\textbf{ETTh1}}} & 96 & .4385 & .4380 & \underline{.4372} & \textbf{.4365} & .4540 & \underline{.4509} & .4519 & \textbf{.4486} & .4402 & \underline{.4378} & .4381 & \textbf{.4364} \\
 & 192 & .5053 & \underline{.5042} & .5048 & \textbf{.5033} & .5146 & \underline{.5114} & .5129 & \textbf{.5095} & .5019 & \underline{.5001} & .5002 & \textbf{.4990} \\
 & 336 & .5689 & \underline{.5686} & .5690 & \textbf{.5678} & .5783 & .5771 & \underline{.5768} & \textbf{.5749} & .5753 & \underline{.5738} & \underline{.5738} & \textbf{.5734} \\
 & 720 & .7714 & \textbf{.7589} & .7711 & \underline{.7644} & .7765 & \textbf{.7606} & .7758 & \underline{.7694} & .7898 & \textbf{.7771} & .7893 & \underline{.7831} \\ \midrule
\multirow{4}{*}{\rotatebox[origin=c]{90}{\textbf{ETTh2}}} & 96 & .2476 & .2463 & \underline{.2459} & \textbf{.2458} & .2348 & \textbf{.2342} & .2346 & \underline{.2343} & .2445 & \underline{.2430} & \textbf{.2418} & \textbf{.2418} \\
 & 192 & .3132 & \underline{.3114} & .3123 & \textbf{.3110} & .3007 & \underline{.3000} & .3008 & \textbf{.2999} & .3006 & \underline{.2971} & \textbf{.2965} & .2977 \\
 & 336 & .3905 & \textbf{.3818} & .3890 & \underline{.3824} & .3737 & \underline{.3706} & .3740 & \textbf{.3691} & .3674 & \underline{.3627} & .3673 & \textbf{.3616} \\
 & 720 & .5599 & \underline{.5508} & .5549 & \textbf{.5365} & .5492 & \underline{.5409} & .5448 & \textbf{.5249} & .5355 & \underline{.5285} & .5319 & \textbf{.5255} \\ \midrule
\multirow{4}{*}{\rotatebox[origin=c]{90}{\textbf{ETTm1}}} & 96 & .3762 & \underline{.3493} & .3507 & \textbf{.3476} & .3708 & \underline{.3545} & .3567 & \textbf{.3495} & .3623 & .3566 & \underline{.3565} & \textbf{.3477} \\
 & 192 & .4376 & .4191 & \underline{.4181} & \textbf{.4130} & .4429 & \underline{.4239} & .4258 & \textbf{.4186} & .4317 & .4237 & \underline{.4219} & \textbf{.4155} \\
 & 336 & .5050 & .4828 & \underline{.4815} & \textbf{.4771} & .5195 & \underline{.4917} & .4921 & \textbf{.4849} & .4959 & .4841 & \underline{.4811} & \textbf{.4766} \\
 & 720 & .5800 & .5656 & \underline{.5630} & \textbf{.5614} & .6010 & .5706 & \underline{.5697} & \textbf{.5668} & .5709 & .5579 & \underline{.5540} & \textbf{.5493} \\ \midrule
\multirow{4}{*}{\rotatebox[origin=c]{90}{\textbf{ETTm2}}} & 96 & .1558 & \underline{.1547} & .1548 & \textbf{.1543} & .1590 & \underline{.1585} & \underline{.1585} & \textbf{.1582} & .1566 & .1557 & \underline{.1552} & \textbf{.1546} \\
 & 192 & .1945 & \textbf{.1938} & .1941 & \underline{.1940} & .1935 & \underline{.1927} & .1929 & \textbf{.1925} & .1922 & \underline{.1913} & .1914 & \textbf{.1907} \\
 & 336 & .2363 & \underline{.2354} & \textbf{.2351} & .2357 & .2342 & \underline{.2327} & \underline{.2327} & \textbf{.2325} & .2330 & .2310 & \underline{.2307} & \textbf{.2304} \\
 & 720 & .3161 & .3105 & \textbf{.3084} & \underline{.3086} & .3121 & .3069 & \underline{.3055} & \textbf{.3046} & .3093 & .3035 & \textbf{.3031} & \underline{.3033} \\ \midrule
\multirow{4}{*}{\rotatebox[origin=c]{90}{\parbox{0.96cm}{\centering\textbf{Exchange}}}} & 96 & .1056 & .1006 & \textbf{.0993} & \underline{.1001} & .0922 & \underline{.0901} & \textbf{.0888} & .0902 & .0918 & .0883 & \textbf{.0871} & \underline{.0881} \\
 & 192 & .2314 & \textbf{.2209} & .2226 & \underline{.2219} & .2057 & .2009 & \underline{.1998} & \textbf{.1974} & .2123 & .2050 & \underline{.2028} & \textbf{.1986} \\
 & 336 & .4071 & \underline{.4013} & .4055 & \textbf{.3592} & .3969 & .3894 & \underline{.3845} & \textbf{.3313} & .4003 & .3920 & \underline{.3898} & \textbf{.3368} \\
 & 720 & 1.1764 & \textbf{.9046} & \underline{1.1078} & 1.1206 & 1.0824 & \underline{.8670} & 1.0187 & \textbf{.8603} & 1.0594 & 1.0301 & \underline{.9844} & \textbf{.8298} \\ \midrule
\multirow{4}{*}{\rotatebox[origin=c]{90}{\textbf{Weather}}} & 96 & .2261 & \underline{.2152} & .2178 & \textbf{.2144} & .2493 & .2346 & \underline{.2343} & \textbf{.2239} & .2360 & \underline{.2241} & .2256 & \textbf{.2193} \\
 & 192 & .2894 & \underline{.2767} & .2793 & \textbf{.2720} & .3056 & \underline{.2883} & .2929 & \textbf{.2811} & .2944 & \underline{.2794} & .2838 & \textbf{.2738} \\
 & 336 & .3534 & \underline{.3332} & .3334 & \textbf{.3281} & .3649 & \underline{.3454} & .3472 & \textbf{.3379} & .3541 & \underline{.3353} & .3371 & \textbf{.3297} \\
 & 720 & .4457 & \underline{.4235} & .4257 & \textbf{.4159} & .4520 & \underline{.4300} & .4339 & \textbf{.4210} & .4428 & \underline{.4234} & .4285 & \textbf{.4151} \\ \midrule
\multicolumn{2}{c}{\textbf{Average}} & .4097 & \textbf{.3895} & .3992 & \underline{.3946} & .4068 & \underline{.3885} & .3961 & \textbf{.3825} & .3999 & .3917 & \underline{.3905} & \textbf{.3782} \\
\midrule
\multicolumn{2}{c}{\textbf{Avg. Rank}} & 3.96 & \underline{2.08} & 2.54 & \textbf{1.42} & 3.92 & \underline{2.21} & 2.71 & \textbf{1.17} & 4.00 & 2.48 & \underline{2.29} & \textbf{1.23} \\
\bottomrule
\end{tabular}
}
}
\end{table}

\subsection{Computational Efficiency}
Table~\ref{tab:efficiency} summarizes the computational efficiency of different TSF-TTA methods using iTransformer as the backbone forecaster model. Overall, the three methods show broadly comparable computational costs in terms of memory usage, throughput, adaptation time, and inference time, suggesting that test-time adaptation adds only modest practical overhead. These results are averaged over all six benchmark datasets and four forecasting horizons. Although DBCM requires a slightly longer adaptation time, the difference is only 0.76~ms compared with PETSA and 2.30~ms compared with TAFAS, while its inference time remains comparable and its throughput is the highest. Moreover, since DBCM is updated only at the adaptation points determined by PAAS~\citep{kim2025battling} rather than at every time step, this millisecond-level overhead is unlikely to become a practical bottleneck, given that the benchmark sampling intervals are several minutes or longer. The more meaningful differences arise in the trade-off between parameter efficiency and predictive performance.

PETSA is the most parameter-efficient method, with the smallest parameter ratio, but its predictive performance is limited, resulting in the highest average MSE among the compared methods. Compared with PETSA, TAFAS achieves better predictive performance, but at the cost of a substantially larger parameter ratio. In contrast, our method achieves the best predictive performance while maintaining a substantially smaller parameter ratio than TAFAS. Since the remaining efficiency metrics stay largely comparable across methods, these results indicate that our method provides a more favorable performance--efficiency trade-off, especially when parameter efficiency is taken into account.

\begin{table}[!ht]
\centering
\renewcommand{\arraystretch}{0.85}
\caption{\textbf{Computational Efficiency Comparison.} Summary of adapter parameter ratio, memory usage, throughput, adaptation time, inference time, and average MSE across TSF-TTA methods. Param. Ratio denotes the number of additional adapter parameters relative to the number of parameters of the source forecaster.}
\small
\label{tab:efficiency}
\resizebox{\textwidth}{!}{%
\begin{tabular}{lcccccc}
\toprule
Method & Param. Ratio (\%) & Peak mem (MB) & Samples/sec & Adapt. time (ms) & Infer. time (ms) & Average MSE\\
\midrule
TAFAS & 16.00 & 512.04 & 12022.58 & 12.93 & 8.30 & 0.3531 \\
PETSA & 1.14 & 491.04 & 11814.63 & 14.47 & 8.89 & 0.3556 \\
DBCM & 5.63 & 492.78 & 12339.85 & 15.23 & 8.77 & 0.3444 \\
\bottomrule
\end{tabular}
}
\end{table}

\subsection{Computational Efficiency in Large-Scale Settings}
\label{app:efficiency_large}
Since the channel-wise LoRA branch and the channel mixer scale with the number of channels, we further evaluate the computational efficiency on the high-dimensional Electricity dataset and on Chronos~\citep{ansari2024chronos}.
Tables~\ref{tab:efficiency_electricity} and~\ref{tab:efficiency_chronos} report the results, where the parameter ratio is defined as in Table~\ref{tab:efficiency}.
The Electricity results are averaged over the four forecasting horizons with iTransformer as the source forecaster, and the Chronos results are averaged over the four ETT datasets with three random seeds.

On Electricity, which contains 321 channels, DBCM achieves the lowest adaptation and inference latency and the highest throughput, while also attaining the lowest MSE.
Its parameter ratio is substantially smaller than that of TAFAS and only slightly higher than that of PETSA, while its peak memory usage remains similar to that of TAFAS.
On Chronos, the adapter parameters account for less than 0.1\% of the forecaster parameters for all methods, and the computational cost is dominated by the frozen foundation model.
Accordingly, DBCM exhibits memory usage and latency comparable to those of TAFAS, lower latency than PETSA, and the highest throughput, while achieving the lowest MSE.

Regarding the scalability with respect to the number of channels \(C\), the channel-wise LoRA branch scales linearly with \(C\), as in PETSA, while the shared adapter and the nonlinear gate are shared across channels.
In contrast, TAFAS assigns a full linear adapter to every channel.
Overall, the number of trainable parameters of DBCM scales as \(\mathcal{O}(L^2 + H^2 + Cr(L+H) + C^2)\), where the \(C^2\) term stems from the channel mixer, which remains substantially smaller than the \(\mathcal{O}(C(L^2 + H^2))\) parameters of TAFAS whenever \(C\) is smaller than \(L^2 + H^2\), as in all evaluated settings.
Although this term increases the parameter count as \(C\) grows, it does not noticeably increase the other computational costs: on Electricity with 321 channels, DBCM still achieves the lowest latency and the highest throughput with memory usage comparable to TAFAS (Table~\ref{tab:efficiency_electricity}).
Together with Table~\ref{tab:efficiency}, these results indicate that DBCM does not incur a consistent latency disadvantage across settings.

\begin{table}[!ht]
\centering
\renewcommand{\arraystretch}{0.85}
\caption{\textbf{Computational Efficiency on Electricity.} Results are averaged over the four forecasting horizons for the iTransformer forecaster.}
\small
\label{tab:efficiency_electricity}
\resizebox{\textwidth}{!}{%
\begin{tabular}{lcccccc}
\toprule
Method & Param. Ratio (\%) & Peak mem (MB) & Samples/sec & Adapt. time (ms) & Infer. time (ms) & Average MSE\\
\midrule
TAFAS & 74.98 & 2109.64 & 2594.97 & 28.64 & 10.05 & 0.1770 \\
PETSA & 25.04 & 1672.08 & 2567.59 & 26.77 & 10.28 & 0.1799 \\
DBCM & 28.80 & 2143.44 & 2710.95 & 26.09 & 9.34 & 0.1744 \\
\bottomrule
\end{tabular}
}
\end{table}

\begin{table}[!ht]
\centering
\renewcommand{\arraystretch}{0.85}
\caption{\textbf{Computational Efficiency on Chronos.} Results are averaged over ETTh1, ETTh2, ETTm1, and ETTm2 with three random seeds.}
\small
\label{tab:efficiency_chronos}
\resizebox{\textwidth}{!}{%
\begin{tabular}{lcccccc}
\toprule
Method & Param. Ratio (\%) & Peak mem (MB) & Samples/sec & Adapt. time (ms) & Infer. time (ms) & Average MSE\\
\midrule
TAFAS & 0.0635 & 4604.39 & 669.64 & 163.85 & 90.68 & 0.3349 \\
PETSA & 0.0126 & 4603.04 & 664.87 & 172.88 & 93.09 & 0.3504 \\
DBCM & 0.0286 & 4604.26 & 670.41 & 164.92 & 90.91 & 0.3176 \\
\bottomrule
\end{tabular}
}
\end{table}

\FloatBarrier
\subsection{Ablation of Input- and Output-Side DBCM}
\label{app:input_output}
We investigate the effect of applying DBCM to the input and output sides of the forecaster.
Following the perspective of TAFAS~\citep{kim2025battling}, the input-side DBCM aims to calibrate distribution-shifted test inputs toward the input distribution that the pre-trained forecaster was trained to handle, whereas the output-side DBCM directly adjusts the forecaster's predictions toward the continuously evolving test distribution.
To assess their individual and combined contributions, we compare three variants: applying DBCM only to the input side, applying DBCM only to the output side, and applying DBCM to both sides.

Table~\ref{tab:input_output} shows that both input-side and output-side DBCM consistently improve over the frozen baseline across backbone models.
The output-side DBCM generally provides stronger gains than the input-side variant, suggesting that directly calibrating forecast outputs is particularly effective for correcting test-time residual errors.
However, the input-side DBCM is also beneficial, especially for DLinear, where it achieves a slightly better average MSE than the output-side variant.
This indicates that input-side calibration can still play an important role by reducing the mismatch before the forecaster processes the test window.

Most importantly, combining input- and output-side DBCM achieves the best average performance for all three backbone models.
These results suggest that the two adaptation sides are complementary: input-side DBCM mitigates distribution shift before forecasting, while output-side DBCM further corrects residual prediction errors after forecasting.
Therefore, applying DBCM to both sides provides a more effective calibration strategy than using either side alone.

\begin{table}[!ht]
\centering
\caption{\textbf{Prediction Performance Comparison by Adaptation Side.} Test MSE is reported for three backbone models on six datasets with four prediction horizons. Bold entries indicate the best performance among methods, while underlined entries denote the second-best.}
\label{tab:input_output}
\setlength{\tabcolsep}{3pt}
\renewcommand{\arraystretch}{0.70}
\setlength{\aboverulesep}{1.5pt}
\setlength{\belowrulesep}{1.5pt}
\setlength{\abovetopsep}{1.5pt}
\setlength{\belowbottomsep}{1.5pt}
\setlength{\cmidrulesep}{1.5pt}
\scriptsize
{
\resizebox{\textwidth}{!}{%
\begin{tabular}{c c c c c c c c c c c c c c}
\toprule
\multicolumn{2}{c}{}
& \multicolumn{4}{c}{\textbf{iTransformer}}
& \multicolumn{4}{c}{\textbf{DLinear}}
& \multicolumn{4}{c}{\textbf{FreTS}} \\
\cmidrule(lr){3-6}\cmidrule(lr){7-10}\cmidrule(lr){11-14}
\multicolumn{2}{c}{}
& Baseline & Input & Output & Both
& Baseline & Input & Output & Both
& Baseline & Input & Output & Both \\
\midrule
\multirow{4}{*}{\rotatebox[origin=c]{90}{\textbf{ETTh1}}} & 96  & .4349 & .4326 & \textbf{.4223} & \underline{.4228} & .4510 & .4379 & \textbf{.4288} & \underline{.4318} & .4360 & \underline{.4315} & \textbf{.4242} & \textbf{.4242} \\
& 192 & .4937 & \underline{.4890} & \textbf{.4805} & \textbf{.4805} & .5042 & .4888 & \textbf{.4828} & \underline{.4835} & .4886 & .4850 & \textbf{.4760} & \underline{.4764} \\
& 336 & .5406 & \textbf{.5320} & .5326 & \underline{.5322} & .5508 & .5405 & \textbf{.5372} & \underline{.5373} & .5466 & .5405 & \textbf{.5353} & \underline{.5354} \\
& 720 & .6914 & .6886 & \underline{.6321} & \textbf{.6298} & .7019 & .6716 & \textbf{.6484} & \underline{.6487} & .7145 & \underline{.7055} & \textbf{.6565} & \textbf{.6565} \\
\midrule
\multirow{4}{*}{\rotatebox[origin=c]{90}{\textbf{ETTh2}}} & 96  & .2430 & .2407 & \textbf{.2378} & \underline{.2388} & .2294 & \textbf{.2260} & .2276 & \underline{.2268} & .2376 & \textbf{.2346} & .2355 & \underline{.2347} \\
& 192 & .2973 & .2938 & \textbf{.2908} & \underline{.2914} & .2835 & \textbf{.2789} & .2801 & \underline{.2794} & .2838 & .2768 & \underline{.2762} & \textbf{.2761} \\
& 336 & .3448 & .3318 & \textbf{.3299} & \underline{.3301} & .3250 & \textbf{.3123} & .3161 & \underline{.3152} & .3184 & .3104 & \underline{.3050} & \textbf{.3048} \\
& 720 & .4258 & .4179 & \textbf{.3937} & \underline{.3939} & .4149 & .3897 & \underline{.3826} & \textbf{.3820} & .3951 & .3846 & \underline{.3654} & \textbf{.3653} \\
\midrule
\multirow{4}{*}{\rotatebox[origin=c]{90}{\textbf{ETTm1}}} & 96  & .3763 & .3542 & \underline{.3462} & \textbf{.3415} & .3705 & .3428 & \underline{.3422} & \textbf{.3420} & .3618 & .3526 & \underline{.3432} & \textbf{.3420} \\
& 192 & .4371 & .4246 & \underline{.4077} & \textbf{.4038} & .4420 & .4080 & \textbf{.4044} & \underline{.4056} & .4304 & .4237 & \textbf{.4059} & \underline{.4060} \\
& 336 & .5029 & .4906 & \underline{.4640} & \textbf{.4628} & .5172 & .4664 & \textbf{.4617} & \underline{.4626} & .4928 & .4908 & \underline{.4625} & \textbf{.4618} \\
& 720 & .5713 & .5641 & \underline{.5285} & \textbf{.5276} & .5927 & \textbf{.5288} & \underline{.5317} & .5321 & .5609 & .5598 & \underline{.5277} & \textbf{.5274} \\
\midrule
\multirow{4}{*}{\rotatebox[origin=c]{90}{\textbf{ETTm2}}} & 96  & .1555 & .1546 & \underline{.1539} & \textbf{.1536} & .1589 & .1577 & \underline{.1575} & \textbf{.1573} & .1563 & .1547 & \underline{.1542} & \textbf{.1536} \\
& 192 & .1938 & .1930 & \textbf{.1915} & \underline{.1923} & .1929 & \underline{.1901} & \textbf{.1899} & .1907 & .1914 & .1905 & \underline{.1883} & \textbf{.1880} \\
& 336 & .2345 & .2339 & \textbf{.2309} & \underline{.2312} & .2325 & \textbf{.2274} & .2290 & \underline{.2289} & .2310 & .2310 & \underline{.2260} & \textbf{.2259} \\
& 720 & .3101 & .3088 & \textbf{.2989} & \underline{.2990} & .3061 & \textbf{.2923} & .2962 & \underline{.2956} & .3026 & .3030 & \textbf{.2924} & \underline{.2930} \\
\midrule
\multirow{4}{*}{\rotatebox[origin=c]{90}{\parbox{0.96cm}{\centering\textbf{Exchange}}}} & 96  & .0916 & \textbf{.0854} & .0868 & \underline{.0858} & .0811 & .0792 & \textbf{.0787} & \underline{.0789} & .0827 & .0798 & \textbf{.0789} & \underline{.0794} \\
& 192 & .1914 & .1799 & \textbf{.1758} & \underline{.1770} & .1708 & .1633 & \underline{.1628} & \textbf{.1626} & .1787 & \underline{.1664} & \underline{.1664} & \textbf{.1659} \\
& 336 & .3375 & .3027 & \textbf{.2826} & \underline{.2844} & .3205 & \underline{.2680} & .2701 & \textbf{.2642} & .3265 & .2907 & \underline{.2723} & \textbf{.2659} \\
& 720 & .9433 & \underline{.8283} & .9424 & \textbf{.8182} & .8383 & \underline{.8016} & .8378 & \textbf{.7066} & .8251 & \underline{.7853} & .8247 & \textbf{.6700} \\
\midrule
\multirow{4}{*}{\rotatebox[origin=c]{90}{\textbf{Weather}}} & 96  & .1719 & .1697 & \textbf{.1620} & \underline{.1630} & .1936 & \underline{.1745} & .1797 & \textbf{.1732} & .1809 & \underline{.1699} & .1716 & \textbf{.1677} \\
& 192 & .2233 & .2170 & \underline{.2108} & \textbf{.2096} & .2383 & .2213 & \underline{.2211} & \textbf{.2194} & .2266 & \underline{.2121} & .2130 & \textbf{.2101} \\
& 336 & .2800 & .2733 & \underline{.2611} & \textbf{.2608} & .2913 & .2703 & \underline{.2702} & \textbf{.2697} & .2806 & .2646 & \underline{.2612} & \textbf{.2602} \\
& 720 & .3576 & .3495 & \textbf{.3348} & \underline{.3360} & .3643 & \textbf{.3305} & .3400 & \underline{.3391} & .3557 & .3362 & \underline{.3339} & \textbf{.3338} \\
\midrule
\multicolumn{2}{c}{\textbf{Average}}
& .3687 & .3565 & \underline{.3499} & \textbf{.3444}
& .3655 & \underline{.3445} & .3449 & \textbf{.3389}
& .3585 & .3492 & \underline{.3415} & \textbf{.3343} \\
\bottomrule
\end{tabular}
}
}
\end{table}

\subsection{Ablation on Channel Mixing}

Table~\ref{tab:channel_mixing} compares four channel mixing strategies for aggregating cross-channel information during adaptation.
The linear channel mixer, used as the default in DBCM, applies a linear transformation along the channel dimension.
MLP extends this design with a two-layer feed-forward network across channels at each time step, using a hidden dimension of 64 and GELU activation to enable learned nonlinear interactions.
Attention represents each channel by its temporal window and computes input-dependent mixing weights between channels, with the projection dimension set to 64.
Uniform performs averaging across channels and broadcasts the shared signal back to each channel.
Overall, the performance differences across channel mixing strategies are small, indicating that DBCM is not highly sensitive to this design choice.
While the MLP mixer obtains marginally lower MSE than Linear on ETTh2, ETTm2, and Exchange Rate, these gains are marginal and do not consistently transfer across all datasets.
The attention-based mixer does not provide clear advantages over Linear, suggesting that input-dependent pairwise channel weighting may introduce unnecessary flexibility for the adaptation signals considered here.
Uniform mixing remains competitive in several settings, including Electricity, but is generally less effective than Linear, indicating that simple channel averaging may be too restrictive to consistently capture useful cross-channel structure.
These results support the use of a simple learnable linear mixer in DBCM, as it provides comparable performance without the additional complexity of nonlinear alternatives.

\begin{table}[!ht]
\centering
\caption{
\textbf{Ablation on Channel Mixing.}
Test MSE across datasets under different channel mixing strategies. Bold entries indicate the best performance among the compared variants.
}
\label{tab:channel_mixing}
\setlength{\tabcolsep}{10pt}
\renewcommand{\arraystretch}{0.75}
\small
\resizebox{\textwidth}{!}{%
\begin{tabular}{lccccccc}
\toprule
 & ETTh1 & ETTh2 & ETTm1 & ETTm2 & Exchange & Weather & Electricity \\
\midrule
Linear & \textbf{.5163} & .3135 & \textbf{.4339} & .2190 & .3413 & \textbf{.2424} & \textbf{.1744} \\
MLP & .5182 & \textbf{.3124} & .4381 & \textbf{.2182} & \textbf{.3410} & \textbf{.2424} & .1756 \\
Attention & .5278 & .3184 & .4458 & .2215 & .3767 & .2449 & .1753 \\
Uniform & .5212 & .3146 & .4431 & .2213 & .3726 & .2453 & .1746\\
\bottomrule
\end{tabular}
}
\end{table}

\subsection{Ablation on the Nonlinear Gating Function}

Table~\ref{tab:nonlinear} compares different nonlinear gating functions used in DBCM.
Overall, Sigmoid achieves the lowest MSE on all datasets except Electricity, where Tanh is marginally better (.1743 vs. .1744), indicating that Sigmoid provides the most stable gating behavior for the proposed calibration branch.
ReLU, SiLU, and Tanh all lead to higher errors than Sigmoid, suggesting that alternative nonlinearities do not provide a consistent advantage in this setting.
In particular, SiLU does not improve performance despite its smooth nonlinear form, while ReLU may be less suitable for modulating correction values due to its unbounded positive activation.
Tanh achieves the best result on Electricity by a very small margin and performs better than ReLU or SiLU on some datasets, but it is generally less effective than Sigmoid across the remaining datasets.
Based on these results, we use Sigmoid as the default nonlinear gating function in DBCM.

\begin{table}[!ht]
\centering
\caption{
\textbf{Ablation on Nonlinear Gating Function.}
Test MSE across datasets under different nonlinear gating functions. Bold entries indicate the best performance among the compared variants.
}
\label{tab:nonlinear}
\setlength{\tabcolsep}{10pt}
\renewcommand{\arraystretch}{0.75}
\small
\resizebox{\textwidth}{!}{%
\begin{tabular}{lccccccc}
\toprule
 & ETTh1 & ETTh2 & ETTm1 & ETTm2 & Exchange & Weather & Electricity \\
\midrule
Sigmoid & \textbf{.5163} & \textbf{.3135} & \textbf{.4339} & \textbf{.2190} & \textbf{.3413} & \textbf{.2424} & .1744 \\
ReLU    & .5278 & .3148 & .4435 & .2204 & .3692 & .2458 & .1756 \\
SiLU    & .5323 & .3154 & .4450 & .2207 & .3718 & .2454 & .1748 \\
Tanh    & .5200 & .3150 & .4400 & .2205 & .3631 & .2462 & \textbf{.1743} \\
\bottomrule
\end{tabular}
}
\end{table}

\subsection{Ablation on Bias Parameterization}

Table~\ref{tab:bias_parameterization} evaluates the effect of different bias parameterizations.
Overall, the performance differences across variants are small, indicating that DBCM is not highly sensitive to this design choice.
We therefore use the channel-wise bias parameterization as the default, as it provides a simple per-channel offset while achieving performance comparable to the other variants.
The full, time-wise, scalar, and no-bias variants yield similar results across datasets.
These results suggest that the main performance gains of DBCM do not depend on a specific bias parameterization.

\begin{table}[!ht]
\centering
\caption{
\textbf{Ablation on Bias Parameterization.}
Test MSE under different additive bias parameterizations.
Here, \(L\) and \(C\) denote the temporal length and number of channels, respectively.
Bold entries indicate the best performance among the compared variants.
}
\label{tab:bias_parameterization}
\setlength{\tabcolsep}{10pt}
\renewcommand{\arraystretch}{0.75}
\resizebox{\textwidth}{!}{%
\begin{tabular}{lccccccc}
\toprule
Bias parameterization & ETTh1 & ETTh2 & ETTm1 & ETTm2 & Exchange & Weather & Electricity \\
\midrule
Full (\(\mathbb{R}^{L \times C}\)) & .5166 & .3136 & .4340 & \textbf{.2190} & \textbf{.3412} & \textbf{.2422} & \textbf{.1744} \\
Channel-wise (\(\mathbb{R}^{C}\)) & \textbf{.5163} & \textbf{.3135} & \textbf{.4339} & \textbf{.2190} & .3413 & .2424 & \textbf{.1744} \\
Time-wise (\(\mathbb{R}^{L}\)) & .5168 & .3136 & .4344 & \textbf{.2190} & .3413 & .2423 & \textbf{.1744} \\
Scalar (\(\mathbb{R}\)) & .5167 & .3136 & .4344 & \textbf{.2190} & .3414 & .2424 & .1745 \\
None & .5168 & .3136 & .4345 & \textbf{.2190} & .3414 & .2424 & \textbf{.1744} \\
\bottomrule
\end{tabular}
}
\end{table}

\begin{sidewaystable}[p]
\centering
\caption{\textbf{Prediction Performance Comparison Across Methods.} Test MSE is reported for six models on six datasets with four prediction horizons. Parenthesized values indicate standard deviation across seeds. Bold entries indicate the best performance among methods, while underlined entries denote the second-best.}
\label{tab:main_std}
\setlength{\tabcolsep}{2.0pt}
\renewcommand{\arraystretch}{0.70}
\setlength{\aboverulesep}{1.2pt}
\setlength{\belowrulesep}{1.2pt}
\setlength{\abovetopsep}{1.2pt}
\setlength{\belowbottomsep}{1.2pt}
\setlength{\cmidrulesep}{1.2pt}
\tiny
\resizebox{\textheight}{!}{%
\begin{tabular}{c c *{24}{c}}
\toprule
\multicolumn{2}{c}{}
& \multicolumn{4}{c}{\textbf{iTransformer}}
& \multicolumn{4}{c}{\textbf{PatchTST}}
& \multicolumn{4}{c}{\textbf{DLinear}}
& \multicolumn{4}{c}{\textbf{OLS}}
& \multicolumn{4}{c}{\textbf{FreTS}}
& \multicolumn{4}{c}{\textbf{MICN}} \\
\cmidrule(lr){3-6}
\cmidrule(lr){7-10}
\cmidrule(lr){11-14}
\cmidrule(lr){15-18}
\cmidrule(lr){19-22}
\cmidrule(lr){23-26}
\multicolumn{2}{c}{}
& Baseline & TAFAS & PETSA & DBCM
& Baseline & TAFAS & PETSA & DBCM
& Baseline & TAFAS & PETSA & DBCM
& Baseline & TAFAS & PETSA & DBCM
& Baseline & TAFAS & PETSA & DBCM
& Baseline & TAFAS & PETSA & DBCM \\
\midrule

\multirow{4}{*}{\rotatebox[origin=c]{90}{\textbf{ETTh1}}}
& 96
& .4349 & .4308 (.0000) & \underline{.4284 (.0004)} & \textbf{.4228 (.0002)}
& .4294 & \underline{.4232 (.0000)} & .4255 (.0001) & \textbf{.4194 (.0001)}
& .4510 & .4425 (.0000) & \underline{.4400 (.0004)} & \textbf{.4318 (.0004)}
& .4511 & \underline{.4420 (.0000)} & .4436 (.0003) & \textbf{.4321 (.0001)}
& .4360 & \underline{.4295 (.0000)} & .4296 (.0003) & \textbf{.4242 (.0001)}
& .4573 & .4471 (.0000) & \underline{.4453 (.0001)} & \textbf{.4342 (.0006)} \\
& 192
& .4937 & .4894 (.0001) & \underline{.4886 (.0003)} & \textbf{.4805 (.0001)}
& .4904 & \underline{.4805 (.0001)} & .4849 (.0001) & \textbf{.4757 (.0001)}
& .5042 & .4946 (.0000) & \underline{.4943 (.0001)} & \textbf{.4835 (.0001)}
& .5044 & \underline{.4923 (.0000)} & .5040 (.0001) & \textbf{.4836 (.0001)}
& .4886 & \underline{.4819 (.0000)} & .4842 (.0002) & \textbf{.4764 (.0001)}
& .5145 & \underline{.4996 (.0000)} & .5063 (.0003) & \textbf{.4900 (.0003)} \\
& 336
& .5406 & \underline{.5382 (.0001)} & .5392 (.0002) & \textbf{.5322 (.0001)}
& .5605 & \underline{.5490 (.0000)} & .5600 (.0001) & \textbf{.5410 (.0003)}
& .5508 & \underline{.5454 (.0000)} & .5455 (.0003) & \textbf{.5373 (.0001)}
& .5510 & \underline{.5415 (.0000)} & .5505 (.0000) & \textbf{.5383 (.0001)}
& .5466 & \underline{.5394 (.0000)} & .5462 (.0001) & \textbf{.5354 (.0002)}
& .5715 & \underline{.5596 (.0000)} & .5708 (.0001) & \textbf{.5553 (.0002)} \\
& 720
& .6914 & \underline{.6520 (.0004)} & .6535 (.0015) & \textbf{.6298 (.0004)}
& .6904 & \underline{.6643 (.0001)} & .6878 (.0002) & \textbf{.6417 (.0006)}
& .7019 & .6713 (.0000) & \underline{.6617 (.0003)} & \textbf{.6487 (.0003)}
& .6996 & \underline{.6654 (.0000)} & .6728 (.0004) & \textbf{.6514 (.0002)}
& .7145 & .6839 (.0000) & \underline{.6795 (.0007)} & \textbf{.6565 (.0002)}
& .7224 & \underline{.6916 (.0001)} & .7201 (.0002) & \textbf{.6743 (.0004)} \\
\midrule

\multirow{4}{*}{\rotatebox[origin=c]{90}{\textbf{ETTh2}}}
& 96
& .2430 & .2406 (.0000) & \underline{.2403 (.0003)} & \textbf{.2388 (.0001)}
& .2309 & \underline{.2298 (.0000)} & .2308 (.0000) & \textbf{.2291 (.0001)}
& .2294 & \underline{.2274 (.0000)} & .2280 (.0001) & \textbf{.2268 (.0001)}
& .2305 & \underline{.2283 (.0000)} & .2286 (.0000) & \textbf{.2279 (.0001)}
& .2376 & \textbf{.2347 (.0000)} & \underline{.2359 (.0001)} & \textbf{.2347 (.0005)}
& .2346 & \underline{.2326 (.0000)} & .2333 (.0001) & \textbf{.2321 (.0001)} \\
& 192
& .2973 & .2931 (.0000) & \underline{.2918 (.0001)} & \textbf{.2914 (.0001)}
& .2871 & \underline{.2802 (.0000)} & .2867 (.0000) & \textbf{.2785 (.0001)}
& .2835 & \textbf{.2793 (.0000)} & .2840 (.0007) & \underline{.2794 (.0001)}
& .2838 & \underline{.2817 (.0000)} & .2822 (.0001) & \textbf{.2802 (.0001)}
& .2838 & \underline{.2776 (.0000)} & .2805 (.0001) & \textbf{.2761 (.0005)}
& .2934 & \underline{.2868 (.0000)} & .2898 (.0001) & \textbf{.2848 (.0002)} \\
& 336
& .3448 & \underline{.3328 (.0001)} & .3334 (.0001) & \textbf{.3301 (.0002)}
& .3311 & \underline{.3163 (.0000)} & .3303 (.0001) & \textbf{.3157 (.0002)}
& .3250 & \underline{.3174 (.0000)} & .3185 (.0007) & \textbf{.3152 (.0002)}
& .3257 & \underline{.3186 (.0000)} & .3251 (.0001) & \textbf{.3157 (.0002)}
& .3184 & \underline{.3051 (.0000)} & .3117 (.0002) & \textbf{.3048 (.0001)}
& .3380 & \underline{.3254 (.0000)} & .3309 (.0001) & \textbf{.3250 (.0003)} \\
& 720
& .4258 & .3999 (.0001) & \underline{.3987 (.0002)} & \textbf{.3939 (.0003)}
& .4067 & \underline{.3886 (.0001)} & .3958 (.0002) & \textbf{.3769 (.0003)}
& .4149 & .3913 (.0000) & \underline{.3878 (.0009)} & \textbf{.3820 (.0004)}
& .4164 & .3902 (.0000) & \underline{.3883 (.0001)} & \textbf{.3831 (.0001)}
& .3951 & \underline{.3690 (.0000)} & .3812 (.0003) & \textbf{.3653 (.0002)}
& .4328 & \underline{.4126 (.0001)} & .4191 (.0003) & \textbf{.4090 (.0002)} \\
\midrule

\multirow{4}{*}{\rotatebox[origin=c]{90}{\textbf{ETTm1}}}
& 96
& .3763 & .3606 (.0001) & \underline{.3601 (.0007)} & \textbf{.3415 (.0004)}
& .3869 & \underline{.3723 (.0000)} & .3829 (.0001) & \textbf{.3469 (.0003)}
& .3705 & \underline{.3496 (.0000)} & .3524 (.0002) & \textbf{.3420 (.0000)}
& .3708 & \underline{.3520 (.0000)} & .3574 (.0004) & \textbf{.3420 (.0002)}
& .3618 & \underline{.3524 (.0000)} & .3525 (.0005) & \textbf{.3420 (.0002)}
& .4063 & \underline{.3723 (.0007)} & .3756 (.0002) & \textbf{.3576 (.0002)} \\
& 192
& .4371 & .4184 (.0001) & \underline{.4145 (.0001)} & \textbf{.4038 (.0001)}
& .4387 & \underline{.4220 (.0000)} & .4328 (.0001) & \textbf{.4057 (.0003)}
& .4420 & \underline{.4159 (.0000)} & .4175 (.0006) & \textbf{.4056 (.0001)}
& .4437 & \underline{.4165 (.0000)} & .4230 (.0002) & \textbf{.4054 (.0001)}
& .4304 & .4177 (.0000) & \underline{.4168 (.0002)} & \textbf{.4060 (.0002)}
& .4479 & .4315 (.0000) & \underline{.4308 (.0002)} & \textbf{.4155 (.0004)} \\
& 336
& .5029 & .4778 (.0002) & \underline{.4739 (.0003)} & \textbf{.4628 (.0001)}
& .4957 & \underline{.4780 (.0001)} & .4939 (.0001) & \textbf{.4593 (.0002)}
& .5172 & .4796 (.0000) & \underline{.4795 (.0005)} & \textbf{.4626 (.0001)}
& .5176 & \underline{.4786 (.0000)} & .4853 (.0005) & \textbf{.4623 (.0001)}
& .4928 & .4752 (.0000) & \underline{.4719 (.0002)} & \textbf{.4618 (.0001)}
& .5154 & \underline{.4885 (.0001)} & .4889 (.0002) & \textbf{.4676 (.0001)} \\
& 720
& .5713 & .5484 (.0000) & \underline{.5472 (.0003)} & \textbf{.5276 (.0001)}
& .5666 & \underline{.5390 (.0001)} & .5615 (.0000) & \textbf{.5255 (.0002)}
& .5927 & \underline{.5482 (.0000)} & .5526 (.0002) & \textbf{.5321 (.0002)}
& .5918 & \underline{.5478 (.0000)} & .5519 (.0003) & \textbf{.5317 (.0002)}
& .5609 & \underline{.5399 (.0000)} & .5404 (.0001) & \textbf{.5274 (.0001)}
& .5749 & \underline{.5404 (.0001)} & .5689 (.0005) & \textbf{.5206 (.0003)} \\
\midrule

\multirow{4}{*}{\rotatebox[origin=c]{90}{\textbf{ETTm2}}}
& 96
& .1555 & \underline{.1542 (.0000)} & .1543 (.0000) & \textbf{.1536 (.0001)}
& .1554 & \underline{.1542 (.0000)} & .1552 (.0000) & \textbf{.1534 (.0001)}
& .1589 & \underline{.1577 (.0000)} & .1579 (.0001) & \textbf{.1573 (.0001)}
& .1602 & .1593 (.0000) & \underline{.1592 (.0000)} & \textbf{.1585 (.0001)}
& .1563 & \underline{.1547 (.0000)} & .1549 (.0000) & \textbf{.1536 (.0001)}
& .1592 & \underline{.1580 (.0000)} & .1582 (.0001) & \textbf{.1574 (.0001)} \\
& 192
& .1938 & \underline{.1926 (.0001)} & .1928 (.0001) & \textbf{.1923 (.0001)}
& .1937 & \underline{.1924 (.0000)} & .1935 (.0000) & \textbf{.1919 (.0001)}
& .1929 & \underline{.1913 (.0000)} & \underline{.1913 (.0001)} & \textbf{.1907 (.0007)}
& .1935 & \underline{.1921 (.0000)} & \textbf{.1919 (.0001)} & \textbf{.1919 (.0001)}
& .1914 & \underline{.1896 (.0000)} & .1898 (.0001) & \textbf{.1880 (.0005)}
& .1955 & .1927 (.0000) & \underline{.1923 (.0001)} & \textbf{.1914 (.0001)} \\
& 336
& .2345 & .2441 (.0146) & \underline{.2325 (.0003)} & \textbf{.2312 (.0001)}
& .2321 & \underline{.2292 (.0000)} & .2317 (.0000) & \textbf{.2282 (.0001)}
& .2325 & \underline{.2290 (.0000)} & .2293 (.0001) & \textbf{.2289 (.0001)}
& .2331 & \textbf{.2298 (.0000)} & \underline{.2301 (.0001)} & \textbf{.2298 (.0001)}
& .2310 & \underline{.2275 (.0000)} & .2305 (.0000) & \textbf{.2259 (.0001)}
& .2421 & .2364 (.0004) & \textbf{.2355 (.0001)} & \underline{.2357 (.0003)} \\
& 720
& .3101 & \underline{.3012 (.0001)} & .3033 (.0018) & \textbf{.2990 (.0001)}
& .3033 & \textbf{.2924 (.0000)} & .3005 (.0001) & \underline{.2933 (.0002)}
& .3061 & .2967 (.0000) & \underline{.2962 (.0001)} & \textbf{.2956 (.0001)}
& .3066 & .2982 (.0000) & \underline{.2977 (.0001)} & \textbf{.2964 (.0001)}
& .3026 & \textbf{.2924 (.0000)} & .3003 (.0001) & \underline{.2930 (.0001)}
& .3176 & \underline{.3040 (.0001)} & .3049 (.0003) & \textbf{.3023 (.0004)} \\
\midrule

\multirow{4}{*}{\rotatebox[origin=c]{90}{\parbox{0.82cm}{\centering\textbf{Exchange}}}}
& 96
& .0916 & .0883 (.0000) & \underline{.0879 (.0002)} & \textbf{.0858 (.0002)}
& .0846 & \underline{.0811 (.0000)} & .0839 (.0001) & \textbf{.0806 (.0003)}
& .0811 & \textbf{.0789 (.0000)} & \underline{.0793 (.0001)} & \textbf{.0789 (.0002)}
& .0814 & \underline{.0795 (.0000)} & .0808 (.0000) & \textbf{.0793 (.0004)}
& .0827 & \textbf{.0784 (.0000)} & .0815 (.0001) & \underline{.0794 (.0002)}
& .0829 & \underline{.0804 (.0000)} & .0818 (.0001) & \textbf{.0800 (.0001)} \\
& 192
& .1914 & \underline{.1834 (.0001)} & .1867 (.0001) & \textbf{.1770 (.0004)}
& .1805 & \underline{.1697 (.0000)} & .1787 (.0001) & \textbf{.1683 (.0006)}
& .1708 & .1643 (.0000) & \underline{.1641 (.0003)} & \textbf{.1626 (.0007)}
& .1723 & \underline{.1640 (.0000)} & .1700 (.0001) & \textbf{.1631 (.0017)}
& .1787 & \underline{.1667 (.0000)} & .1765 (.0000) & \textbf{.1659 (.0005)}
& .1886 & \underline{.1771 (.0000)} & .1853 (.0001) & \textbf{.1742 (.0011)} \\
& 336
& .3375 & \underline{.3146 (.0004)} & .3243 (.0001) & \textbf{.2844 (.0051)}
& .5375 & \underline{.4585 (.0005)} & .4755 (.0003) & \textbf{.4087 (.0312)}
& .3205 & \underline{.2883 (.0000)} & .2931 (.0003) & \textbf{.2642 (.0011)}
& .3226 & \underline{.2851 (.0000)} & .2876 (.0008) & \textbf{.2613 (.0013)}
& .3265 & \underline{.2922 (.0000)} & .3198 (.0004) & \textbf{.2659 (.0012)}
& .3529 & \underline{.3116 (.0000)} & .3315 (.0006) & \textbf{.2829 (.0021)} \\
& 720
& .9433 & \underline{.8388 (.0000)} & .8894 (.0024) & \textbf{.8182 (.0290)}
& .8814 & \underline{.8812 (.0000)} & .8814 (.0000) & \textbf{.6609 (.0837)}
& .8383 & .8309 (.0000) & \underline{.8109 (.0010)} & \textbf{.7066 (.0793)}
& .8360 & \underline{.6485 (.0000)} & .8114 (.0009) & \textbf{.5439 (.0064)}
& .8251 & \underline{.7606 (.0000)} & .8141 (.0001) & \textbf{.6700 (.0115)}
& 1.4040 & \textbf{1.1221 (.0000)} & 1.3355 (.0011) & \underline{1.2209 (.1460)} \\
\midrule

\multirow{4}{*}{\rotatebox[origin=c]{90}{\textbf{Weather}}}
& 96
& .1719 & \underline{.1642 (.0001)} & .1661 (.0007) & \textbf{.1630 (.0003)}
& .1738 & \underline{.1672 (.0000)} & .1730 (.0001) & \textbf{.1657 (.0002)}
& .1936 & .1788 (.0000) & \underline{.1769 (.0005)} & \textbf{.1732 (.0002)}
& .1955 & \underline{.1808 (.0000)} & .1819 (.0015) & \textbf{.1763 (.0004)}
& .1809 & \underline{.1693 (.0000)} & .1713 (.0006) & \textbf{.1677 (.0001)}
& .1720 & \underline{.1699 (.0002)} & .1726 (.0002) & \textbf{.1670 (.0004)} \\
& 192
& .2233 & \underline{.2104 (.0003)} & .2129 (.0001) & \textbf{.2096 (.0001)}
& .2207 & \underline{.2103 (.0000)} & .2183 (.0001) & \textbf{.2073 (.0002)}
& .2383 & \underline{.2236 (.0000)} & .2255 (.0003) & \textbf{.2194 (.0003)}
& .2403 & \underline{.2219 (.0000)} & .2272 (.0001) & \textbf{.2196 (.0003)}
& .2266 & \underline{.2109 (.0000)} & .2179 (.0002) & \textbf{.2101 (.0001)}
& .2222 & \underline{.2155 (.0001)} & .2237 (.0009) & \textbf{.2148 (.0001)} \\
& 336
& .2800 & \underline{.2610 (.0000)} & .2646 (.0001) & \textbf{.2608 (.0001)}
& .2760 & \underline{.2598 (.0001)} & .2721 (.0001) & \textbf{.2586 (.0002)}
& .2913 & \underline{.2714 (.0000)} & .2755 (.0003) & \textbf{.2697 (.0001)}
& .2920 & \underline{.2706 (.0000)} & .2772 (.0001) & \textbf{.2697 (.0002)}
& .2806 & \underline{.2609 (.0000)} & .2675 (.0003) & \textbf{.2602 (.0001)}
& .2747 & \underline{.2626 (.0002)} & .2701 (.0008) & \textbf{.2624 (.0006)} \\
& 720
& .3576 & \underline{.3404 (.0000)} & .3498 (.0001) & \textbf{.3360 (.0006)}
& .3512 & .3644 (.0001) & \underline{.3410 (.0005)} & \textbf{.3315 (.0001)}
& .3643 & \underline{.3490 (.0000)} & .3501 (.0006) & \textbf{.3391 (.0004)}
& .3646 & \underline{.3443 (.0000)} & .3626 (.0001) & \textbf{.3398 (.0003)}
& .3557 & \underline{.3364 (.0000)} & .3536 (.0001) & \textbf{.3338 (.0002)}
& .3491 & \underline{.3371 (.0001)} & .3546 (.0028) & \textbf{.3298 (.0003)} \\
\midrule

\multicolumn{2}{c}{\textbf{Average}}
& .3687 & \underline{.3531} & .3556 & \textbf{.3444}
& .3710 & \underline{.3585} & .3657 & \textbf{.3402}
& .3655 & .3509 & \underline{.3505} & \textbf{.3389}
& .3660 & \underline{.3429} & .3538 & \textbf{.3326}
& .3585 & \underline{.3436} & .3503 & \textbf{.3343}
& .3946 & \underline{.3690} & .3844 & \textbf{.3660} \\
\bottomrule
\end{tabular}
}
\end{sidewaystable}

\section{Detailed Dataset Information}
\label{app:data_description}
We provide detailed information on the benchmark datasets used in our experiments, including their domains, temporal resolutions, and brief descriptions.
Table~\ref{tab:dataset_summary} presents an overview of all datasets. Table~\ref{tab:dataset_stats} summarizes the statistical characteristics of the benchmark datasets, including the number of variables, temporal resolution, and stationarity measured by the Augmented Dickey--Fuller (ADF) statistic~\citep{elliott1996adf}.
The ADF statistic serves as an indicator of non-stationarity, where values closer to zero imply stronger non-stationarity~\citep{kim2025battling}. Although it does not directly measure the discrepancy between training and test distributions, it provides a proxy for how strongly the temporal dynamics evolve over time.
As shown in the table, Exchange exhibits the highest ADF value ($-1.90$), indicating the strongest non-stationarity among the datasets.
In contrast, datasets such as Weather ($-26.68$) show highly negative ADF values, suggesting relatively stable temporal dynamics.
Overall, these datasets span a wide range of non-stationarity.

\begin{table}[!ht]
\centering
\caption{\textbf{Benchmark Dataset Characteristics.}
We summarize the domain, sampling frequency, number of channels, and a brief description of each dataset.}
\label{tab:dataset_summary}
\resizebox{\textwidth}{!}{%
\begin{tabular}{l l l c l}
\toprule
Dataset & Domain & Frequency & \# Channels & Description \\
\midrule
Electricity & Electricity & 1 hour & 321 &
Hourly electricity consumption from 2012 to 2014. \\

Weather & Weather & 10 minutes & 21 &
Meteorological indicators recorded every 10 minutes in 2020. \\

ETTh & Electricity & 1 hour & 7 &
Hourly transformer load and oil-temperature data. \\

ETTm & Electricity & 15 minutes & 7 &
15-minute transformer load and oil-temperature data. \\

Exchange & Finance & 1 day & 8 &
Daily exchange rates from eight countries. \\

\bottomrule
\end{tabular}}
\end{table}

\begin{table}[!ht]
\centering
\caption{\textbf{Benchmark Dataset Statistics.}
We report the number of time steps and ADF test statistic for each dataset.}
\label{tab:dataset_stats}
\begin{tabular}{lcc}
\toprule
Dataset  & Time Steps & ADF \\
\midrule
ETTh1     & 17,420  & -5.91 \\
ETTm1     & 69,680  & -14.98 \\
ETTh2     & 17,420  & -4.13 \\
ETTm2     & 69,680  & -5.66 \\
Exchange  & 7,588   & -1.90 \\
Weather   & 52,696  & -26.68 \\
\bottomrule
\end{tabular}
\end{table}

\section{Training Setup}
\label{app:setup}
We follow the experimental setup of TAFAS~\citep{kim2025battling}. For backbone pre-training, we perform a hyperparameter search over learning rates \(\{1\mathrm{e}{-3}, 1\mathrm{e}{-4}, 1\mathrm{e}{-5}\}\) and weight decay values \(\{0.0, 1\mathrm{e}{-3}, 1\mathrm{e}{-4}\}\).
Each configuration is trained with three random seeds, and we select the hyperparameter setting that achieves the lowest validation MSE.
All backbone models are pre-trained for 30 epochs with a batch size of 64 using the Adam optimizer~\citep{kingma2015adam} and cosine learning rate scheduling~\citep{loshchilov2017sgdr}.
For all TSF-TTA methods, we follow the tuning procedure of TAFAS~\citep{kim2025battling}, including the same data split for hyperparameter selection, and search the learning rate over \(\{5\mathrm{e}{-3}, 3\mathrm{e}{-3}, 1\mathrm{e}{-3}, 5\mathrm{e}{-4}, 1\mathrm{e}{-4}\}\), the weight decay over \(\{0.0, 1\mathrm{e}{-4}\}\), and the initial value of the gating parameter over \(\{0.01, 0.05, 0.1, 0.3\}\), so that all methods receive the same tuning budget and selection procedure.
This tuning procedure is applied consistently across all experimental settings.
For reproducibility, the selected hyperparameter configurations for DBCM will be provided in our official implementation.
All experiments are conducted on a single RTX 4090 GPU, except those involving Moirai~\citep{woo2024moirai}, Chronos~\citep{ansari2024chronos}, and Chronos-2~\citep{ansari2025chronos2}, which are conducted on a single RTX 6000 ADA GPU.

\section{Drift Descriptors}
\label{app:drift_descriptors}
We provide detailed definitions of the channel-wise drift descriptors used to characterize raw data-side drift.
Following the notation in the main text, \(\mathbf{x}_t \in \mathbb{R}^{C}\) denotes the multivariate observation at time \(t\).
For channel-wise descriptor computation, we write \(x_t^{(c)} = (\mathbf{x}_t)_c\) for the scalar value of channel \(c\) at time \(t\).
All descriptors are computed on channel-wise standardized series:
\begin{equation}
\tilde{x}_t^{(c)}
=
\frac{x_t^{(c)}-\mu^{\mathrm{train}}_c}
{\sigma^{\mathrm{train}}_c},
\end{equation}
where \(\mu^{\mathrm{train}}_c\) and \(\sigma^{\mathrm{train}}_c\) denote the mean and standard deviation of channel \(c\) computed over the training split.

We use sliding lookback windows of length \(L=96\) with stride 1.
For a descriptor window ending at time \(\tau\), let
\begin{equation}
\mathbf{w}_{\tau,c}
=
[\tilde{x}_{\tau-L+1}^{(c)}, \ldots, \tilde{x}_{\tau}^{(c)}]
\in \mathbb{R}^{L}.
\end{equation}

We compute six descriptor types, grouped into low-order and frequency/temporal descriptor families.
The low-order descriptors consist of mean, log-variance, and slope, while the frequency/temporal descriptors consist of dominant frequency, spectral-band energy, and autocorrelation.

For the low-order descriptors, we define
\begin{equation}
\mu_{\tau,c}
=
\frac{1}{L}\sum_{i=1}^{L} w_{\tau,c}[i],
\end{equation}
\begin{equation}
v_{\tau,c}
=
\frac{1}{L}\sum_{i=1}^{L}
\left(w_{\tau,c}[i]-\mu_{\tau,c}\right)^2,
\end{equation}
and use the log-variance descriptor
\begin{equation}
\tilde{v}_{\tau,c}
=
\log\left(v_{\tau,c}+\epsilon\right),
\end{equation}
where \(\epsilon>0\) is a small constant for numerical stability.

For the slope descriptor, we use a centered normalized time axis
\begin{equation}
a_i = -\frac{1}{2} + \frac{i-1}{L-1},
\qquad i=1,\ldots,L,
\end{equation}
and compute
\begin{equation}
s_{\tau,c}
=
\frac{\sum_{i=1}^{L} a_i\, w_{\tau,c}[i]}
{\sum_{i=1}^{L} a_i^2}.
\end{equation}

The frequency and temporal descriptors are computed from the centered signal
\begin{equation}
\mathbf{u}_{\tau,c}
=
\mathbf{w}_{\tau,c}-\mu_{\tau,c}\mathbf{1}_{L}.
\end{equation}
Let
\begin{equation}
P_{\tau,c}[f]
=
\left|
\operatorname{FFT}(\mathbf{u}_{\tau,c})[f]
\right|^2,
\qquad
f=0,\ldots,\left\lfloor \frac{L}{2}\right\rfloor,
\end{equation}
denote the one-sided Fourier power spectrum.

We define the dominant-frequency descriptor as the normalized index of the strongest positive-frequency bin:
\begin{equation}
f^{\star}_{\tau,c}
=
\operatorname*{arg\,max}_{1 \le f \le \lfloor L/2 \rfloor} \ P_{\tau,c}[f],
\qquad
\phi_{\tau,c}
=
\frac{f^{\star}_{\tau,c}}{\lfloor L/2 \rfloor}.
\end{equation}

To capture spectral energy, we partition the positive-frequency bins into \(N_{\mathrm{band}}\) disjoint bands
\(\mathcal{B}_1,\ldots,\mathcal{B}_{N_{\mathrm{band}}}\) and compute log band energies
\begin{equation}
e^{(b)}_{\tau,c}
=
\log\left(
\sum_{f \in \mathcal{B}_b} P_{\tau,c}[f] + \epsilon
\right),
\qquad b=1,\ldots,N_{\mathrm{band}}.
\end{equation}
In the default implementation, \(N_{\mathrm{band}}=3\).

For autocorrelation, we use a lag set \(\Lambda=\{\ell_1,\ldots,\ell_{N_{\mathrm{lag}}}\}\) and compute
\begin{equation}
\rho^{(\ell)}_{\tau,c}
=
\frac{
\sum_{i=1}^{L-\ell}
u_{\tau,c}[i]u_{\tau,c}[i+\ell]
}{
\sum_{i=1}^{L}
u_{\tau,c}[i]^2 + \epsilon
}.
\qquad \ell \in \Lambda.
\end{equation}
By default, \(\Lambda\) is chosen as \(\{1, p_{\mathrm{data}}, 2p_{\mathrm{data}}\}\) after removing invalid lags, where \(p_{\mathrm{data}}\) denotes the dataset period length.

The concatenated descriptor vector for channel \(c\) at time \(\tau\) is
\begin{equation}
\mathbf{d}_{\tau,c}
=
[
\mu_{\tau,c},
\tilde{v}_{\tau,c},
s_{\tau,c},
\phi_{\tau,c},
e^{(1)}_{\tau,c}, \ldots, e^{(N_{\mathrm{band}})}_{\tau,c},
\rho^{(\ell_1)}_{\tau,c}, \ldots, \rho^{(\ell_{N_{\mathrm{lag}}})}_{\tau,c}
]
\in \mathbb{R}^{K},
\end{equation}
where
\begin{equation}
K = 4 + N_{\mathrm{band}} + N_{\mathrm{lag}}.
\end{equation}
Under the default setting \(N_{\mathrm{band}}=3\) and \(N_{\mathrm{lag}}=3\), we have \(K=10\).

We consider two drift references.
Under the source reference, each test-window descriptor is compared against the mean descriptor over the training windows:
\begin{equation}
\bar{\mathbf{d}}^{\mathrm{src}}_{c}
=
\frac{1}{|\mathcal{I}_{\mathrm{train}}|}
\sum_{\tau \in \mathcal{I}_{\mathrm{train}}}
\mathbf{d}_{\tau,c},
\end{equation}
where \(\mathcal{I}_{\mathrm{train}}\) denotes the set of descriptor-window endpoints in the training split.
The source-referenced drift is then defined as
\begin{equation}
\Delta^{\mathrm{src}}_{\tau,c}
=
\mathbf{d}_{\tau,c}
-
\bar{\mathbf{d}}^{\mathrm{src}}_{c}.
\end{equation}

Under the previous-window reference, each test-window descriptor is compared against the descriptor of the immediately preceding test window:
\begin{equation}
\Delta^{\mathrm{prev}}_{\tau,c}
=
\mathbf{d}_{\tau,c}
-
\mathbf{d}_{\tau-1,c}.
\end{equation}

For each descriptor family \(j\), the resulting drift tensor has shape \([N_{\mathrm{win}}, C, K_j]\), where \(N_{\mathrm{win}}\) is the number of descriptor windows, \(C\) is the number of channels, and \(K_j\) is the feature dimension of that descriptor family.
We reshape each tensor into a channel-space matrix
\begin{equation}
\mathbf{S}^{(j)} \in \mathbb{R}^{C \times (N_{\mathrm{win}}K_j)}
\end{equation}
for channel-space spectral analysis.
For the combined analysis, we first normalize each descriptor-family matrix and then concatenate them along the feature dimension.

\section{Additional Related Work}
\label{app:related_work}
\paragraph{Time Series Forecasting.}
Recent time series forecasting models have introduced diverse architectures for capturing complex temporal dependencies.
Transformer-based forecasters, such as PatchTST~\citep{nie2022time} and iTransformer~\citep{liu2023itransformer}, model long-range temporal and inter-channel relationships, while alternative approaches explore different inductive biases.
DLinear~\citep{zeng2022arete} demonstrates the effectiveness of simple linear mappings, FreTS~\citep{yi2023frequency} leverages frequency-domain representations, and MICN~\citep{wang2023micn} adopts multi-scale convolutions.
More recently, time series foundation models (TSFMs), including Chronos~\citep{ansari2024chronos, ansari2025chronos2} and TimesFM~\citep{das2024timesfm}, have been proposed to improve generalization through pre-training on large-scale time series corpora.
Despite strong performance, they implicitly assume stationarity, whereas real-world time series exhibit evolving and non-stationary distributions, motivating test-time adaptation methods~\citep{kim2025battling}.

\paragraph{Test-Time Adaptation.}
Test-time adaptation (TTA) aims to improve model performance under distribution shift by adapting a pre-trained model using test-time data.
Representative methods such as Tent~\citep{wang2020tent} adapt models by minimizing prediction entropy, and subsequent works improve stability and efficiency under continual distribution shifts~\citep{wang2022continual, niu2022efficient}.
These methods, however, are primarily designed for classification, where entropy over class probabilities provides an unsupervised adaptation objective.
Extending TTA to time series forecasting requires different assumptions.
Forecasting is a regression task, so entropy minimization objectives are not directly applicable.
Moreover, future values become observable after prediction, and the sequential structure of time series enables partially observed ground truth before the full horizon is available.
At the same time, test windows are temporally dependent rather than IID (Independent and Identically Distributed).
Therefore, TSF-TTA should exploit delayed and partially observed targets while adapting a frozen forecaster in a way that respects the streaming, non-IID nature of time series.

\paragraph{Cross-Channel Dependency in Multivariate Time Series Forecasting.}
Cross-channel dependency is a key factor in multivariate time series forecasting.
Recent works have explored various mechanisms to capture such dependencies during representation learning.
Crossformer~\citep{zhang2023crossformer} introduces cross-dimension attention to explicitly model inter-channel interactions.
TSMixer~\citep{chen2023tsmixer} demonstrates that simple MLP-based alternating time and feature mixing can effectively capture cross-channel structure without relying on attention.
MTS-Mixer~\citep{li2025mts} further extends this line of work by proposing a factorized architecture that separately performs temporal and channel mixing to efficiently model multivariate dependencies.
iTransformer~\citep{liu2023itransformer} revisits the Transformer by treating channels as tokens, enabling direct modeling of cross-channel relationships via self-attention.
TimeXer~\citep{wang2024timexer} enhances this paradigm with more expressive mixing mechanisms across time and channels.
PCD~\citep{lee2026dataset} further argues that channel dependence can be partial and dataset-specific, and incorporates channel masks constructed from channel similarity and learnable domain parameters into Transformer attention matrices.
Finally, CrossLinear~\citep{zhou2025crosslinear} explores linear mappings with cross-channel interactions, showing that even simple linear structures can capture inter-variable dependencies. In contrast, we consider cross-channel structure in the context of test-time adaptation, using channel mixing to provide cross-channel context before calibration.

\paragraph{Cross-Channel Structure of Forecast Residuals.}
A general theoretical characterization of when forecast residuals are low-rank in the channel space requires restrictive assumptions, since residuals depend on both the data-generating process and the fitted forecaster~\citep{west1996asymptotic}.
Nevertheless, multivariate forecast errors can exhibit contemporaneous, temporal, and cross-lag dependencies~\citep{zheng2024multivariate}, and high-dimensional cross-channel dependence can sometimes be represented compactly through low-rank structures~\citep{salinas2019high}.
Such structure may arise when many channels share a small number of factors that are incompletely modeled by the forecaster, such as temperature shocks affecting regional electricity demand.
Conversely, cross-channel correlation in the raw signals does not by itself imply concentrated residuals: if the forecaster captures the shared factors well, the remaining residuals are dominated by channel-specific components.
Consistently, raw data-side drift is less concentrated in the channel space than forecast residuals once it is characterized beyond simple statistical shifts (Appendix~\ref{app:raw_drift}).
These observations support the plausibility of our empirical hypothesis, but do not imply that residual low-rankness is a universal property.
Accordingly, we treat the low-rank structure of forecast residuals as an empirical hypothesis whose strength varies across datasets and forecasters (Section~\ref{sec:analysis}).
This variation motivates the dual-branch design of DBCM, which combines parameters shared across channels with a channel-wise LoRA branch that provides limited channel-specific flexibility.

\paragraph{Online Learning for Time Series Forecasting.}

Handling distribution shift in time series has been widely studied in online learning.
FSNet~\citep{pham2022learning} introduces a dual-timescale framework that balances fast adaptation and long-term stability.
OneNet~\citep{wen2023onenet} adopts an online ensembling strategy to cope with concept drift.
SOLID~\citep{chen2024calibration} explores calibration-based approaches to detect and correct context-driven distribution shifts.
More recently, PROCEED~\citep{zhao2025proactive} proposes a proactive adaptation mechanism that anticipates distribution shift, ELF~\citep{lee2025lightweight} focuses on lightweight online adaptation for time series foundation model forecasts, and DynaME~\citep{hong2026dynamic} dynamically selects multi-period experts for recurring patterns and a general expert for emergent shifts.
These approaches handle distribution shift by continuously updating the forecasting system itself, and do not explicitly model the structure of adaptation signals.
In contrast, our scope is restricted to the TSF-TTA setting, in which the pre-trained source forecaster remains frozen and only external calibration modules are updated.
Within this setting, we show that adaptation signals exhibit compact shared structure across channels and propose a structured calibration framework that decomposes them into shared and residual components.

\end{document}

%% file: math_commands.tex
\usepackage{amsmath,amsfonts,bm}

\def\eqref#1{equation~\ref{#1}}

\def\1{\bm{1}}

\DeclareMathAlphabet{\mathsfit}{\encodingdefault}{\sfdefault}{m}{sl}
\SetMathAlphabet{\mathsfit}{bold}{\encodingdefault}{\sfdefault}{bx}{n}